\documentclass[10pt,twocolumn,letterpaper]{article}
\usepackage[pagenumbers]{cvpr}
\usepackage{indentfirst}
\usepackage{mathtools}
\usepackage{xspace}

\newcommand{\mba}{\mathbf{a}}

\newcommand{\mbe}{\mathbf{e}}

\newcommand{\mbu}{\mathbf{u}}

\newcommand{\mbx}{\mathbf{x}}

\newcommand{\mbn}{\mathbf{n}}

\newcommand{\calA}{\mathcal{A}}

\newcommand{\calE}{\mathcal{E}}

\newcommand{\calG}{\mathcal{G}}

\newcommand{\calL}{\mathcal{L}}

\newcommand{\calW}{\mathcal{W}}

\newcommand{\ignore}[1]{}

\makeatletter
\DeclareRobustCommand\onedot{\futurelet\@let@token\@onedot}
\def\@onedot{\ifx\@let@token.\else.\null\fi\xspace}
\def\eg{{e.g}\onedot} %\def\Eg{{E.g}\onedot}
\def\ie{{i.e}\onedot} %\def\Ie{{I.e}\onedot}
\makeatother

\definecolor{cvprblue}{rgb}{0.21,0.49,0.74}
\definecolor{skyblue}{rgb}{0.53, 0.81, 0.92}
\usepackage[pagebackref,breaklinks,colorlinks,allcolors=cvprblue]{hyperref}
\usepackage{textgreek}
\usepackage[utf8]{inputenc}
\usepackage{tipa}
\usepackage[subtle,mathdisplays=normal,wordspacing=normal]{savetrees}

\newcommand\blfootnote[1]{%
  \begingroup
  \renewcommand\thefootnote{}\footnote{#1}%
  \addtocounter{footnote}{-1}%
  \endgroup
}

\title{EmoTalkingGaussian: Continuous Emotion-conditioned Talking Head Synthesis}

\author{Junuk Cha\textsuperscript{1,2\dag} \qquad Seongro Yoon\textsuperscript{2} \qquad Valeriya Strizhkova\textsuperscript{2} \qquad Francois Bremond\textsuperscript{2} \qquad Seungryul Baek\textsuperscript{1} \vspace{0.3em} \\
{\normalsize $^1$UNIST} \qquad
{\normalsize $^2$Inria}
}

\begin{document}
\twocolumn[{
\maketitle
\begin{center}
    \captionsetup{type=figure}
    \vspace{-6mm}
    \includegraphics[width=0.96\textwidth]{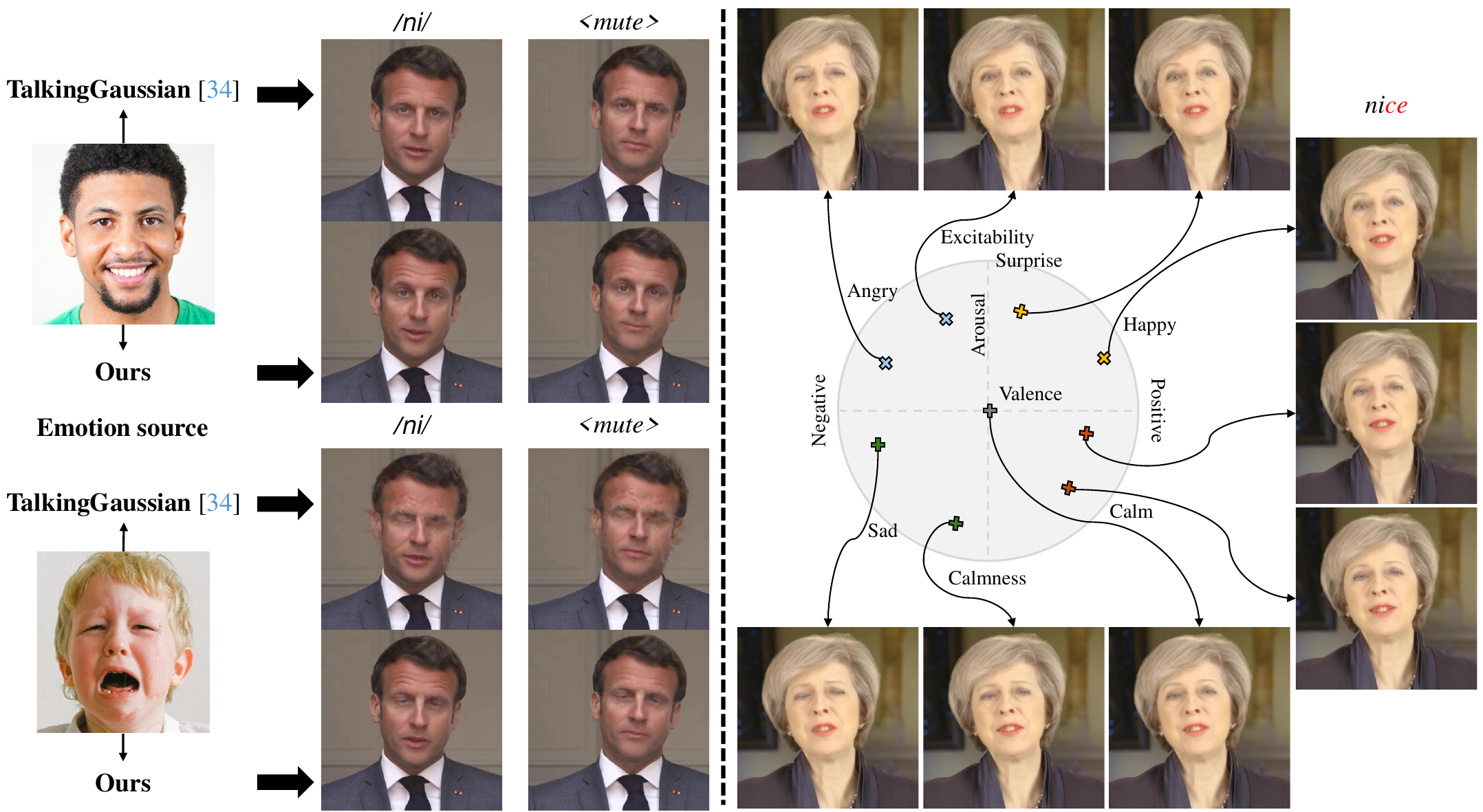}
    %\vspace{-3mm}
    \captionof{figure}{The state-of-the-art 3D talking head synthesis method, TalkingGaussian~\cite{li2024talkinggaussian}, manipulates expressions based on action units~\cite{ekman1978facial}; however, its ability to express diverse emotions is limited, and the image quality becomes inferior when representing unseen emotional expression of the emotion source image~\cite{pexels}. Our method can reflect diverse expressions and emotions based on action units as well as valence/arousal~\cite{russell1980circumplex}, and it renders the talking head with lip shape well-aligned to the input audio (/\textit{ni}/ and \ \textless \textit{mute} \textgreater), as shown in the left panel. The right panel demonstrates our method's capability to convey continuous emotions through valence/arousal adjustments, while keeping the lip synchronized to the audio. The ``ce" in "nice," which the speaker is pronouncing, is highlighted in \textcolor{red}{red}.
    %However, it has limitations in expressing various emotions.
    }
    \label{fig:teaser}
\end{center}
}]

\blfootnote{$\dag$ This research was conducted when Junuk Cha was an intern at Inria.}

\begin{abstract}
3D Gaussian splatting-based talking head synthesis has recently gained attention for its ability to render high-fidelity images with real-time inference speed. However, since it is typically trained on only a short video that lacks the diversity in facial emotions, the resultant talking heads struggle to represent a wide range of emotions. To address this issue, we propose a lip-aligned emotional face generator and leverage it to train our EmoTalkingGaussian model. It is able to manipulate facial emotions conditioned on continuous emotion values (\ie, valence and arousal); while retaining synchronization of lip movements with input audio. Additionally, to achieve the accurate lip synchronization for in-the-wild audio, we introduce a self-supervised learning method that leverages a text-to-speech network and a visual-audio synchronization network. We experiment our EmoTalkingGaussian on publicly available videos and have obtained better results than state-of-the-arts in terms of image quality (measured in PSNR, SSIM, LPIPS), emotion expression (measured in V-RMSE, A-RMSE, V-SA, A-SA, Emotion Accuracy), and lip synchronization (measured in LMD, Sync-E, Sync-C), respectively.
\end{abstract}
\section{Introduction}
\label{sec:intro}
3D Gaussian splatting (3DGS)~\cite{kerbl20233d} has been recently established as the alternative to neural radiance fields (NeRF)~\cite{mildenhall2021nerf}, offering substantial improvements in both rendering speed and quality. Talking head synthesis domain also reflects the trend: NeRF-based approaches~\cite{peng2024synctalk,ye2023geneface,tang2022radnerf,shen2022learning,li2023efficient,guo2021ad,chatziagapi2023lipnerf} have been prevailed; while they are being rapidly replaced by the 3D Gaussian splatting-based approaches~\cite{li2024talkinggaussian,cho2024gaussiantalker,yu2024gaussiantalker,he2024emotalk3d}, thanks to its real-time speed and high-fidelity rendering quality.

Despite the advancements, we argue that existing pipelines lack the important aspect of human \emph{emotions}. We observed that existing models~\cite{li2024talkinggaussian,cho2024gaussiantalker} are able to handle basic facial expressions such as \emph{eye blinking} and \emph{eyebrow movement}, as seen in the 3-5 minute training video; however they struggle to represent continuous and diverse emotions such as \emph{happy}, \emph{sad}, \emph{angry}, etc. When talking, humans convey diverse emotions. To achieve the truly life-like talking heads by filling the gap, we insist that synthesized talking heads need to represent such diverse human emotions while talking. Although He~\etal~\cite{he2024emotalk3d} proposed a method that relies on collecting new data, a significant drawback is the high cost involved, as additional data must be captured to train a 3D Gaussian model for each new person.

\begin{figure}[t]
    \centering
    \includegraphics[width=0.99\linewidth]{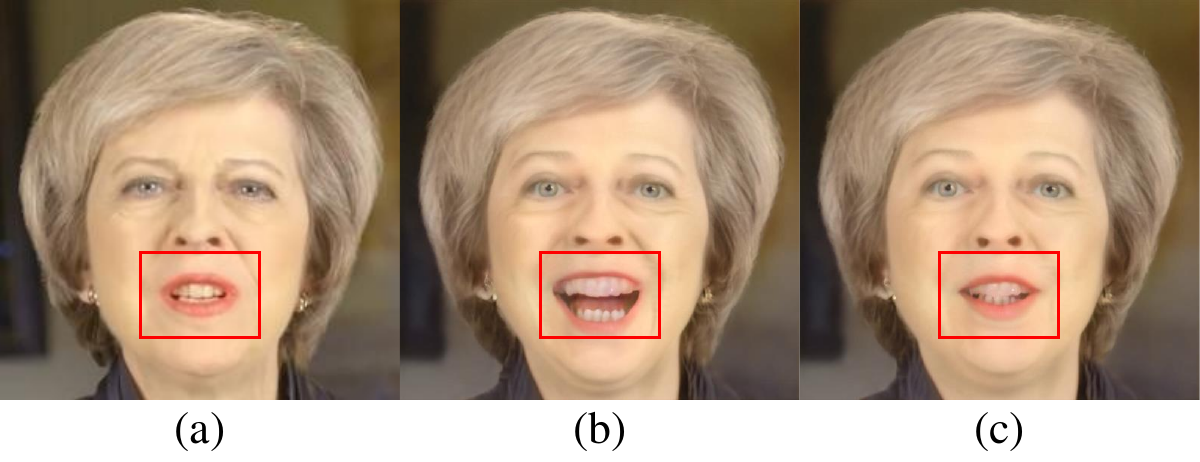}
    \vspace{-3mm}
    \caption{(a) shows the source image, (b) and (c) represent images for the `\emph{happy\&surprise}' emotion (valence of 0.8, arousal of 0.6), which are generated by EmoStyle~\cite{azari2024emostyle} and our lip-aligned emotional face generator, respectively. 
    % EmoStyle~\cite{azari2024emostyle} significantly changes lip regions (in \textcolor{red}{red} boxes); since it only reflects valence/arousal values, without any constraints on lips. Our lip-aligned emotional face generator is able to synthesize faces reflecting valence/arousal values, while retaining lips.
    %Since EmoStyle does not consider the lip alignment, training our talking head synthesis model using the data it generates can cause lip movements to be out of sync with the input audio.
    }
    \label{fig:motivate finetune emotstyle}
    \vspace{-3mm}
\end{figure}

In this paper, we propose EmoTalkingGaussian that integrates continuous emotional expression into 3D Gaussian splatting-based talking head synthesis. To manipulate facial emotions, we utilize valence and arousal~\cite{russell1980circumplex} as conditions of EmoTalkingGaussain. Valence represents the degree of positiveness or negativeness, and arousal indicates the level of excitability or calmness, both ranging from -1 to 1. Unlike action units~\cite{ekman1978facial} that capture basic facial expressions such as \emph{eye blinking}, valence and arousal enable continuous adjustments to facial emotions, (\eg, \emph{happy}, \emph{surprise}, \emph{sad}, etc), as shown in Fig.~\ref{fig:teaser}. To train EmoTalkingGaussian on diverse emotional facial images not seen in the original train video, one solution is to augment the data using EmoStyle~\cite{azari2024emostyle}, which modifies the emotion of a source image based on valence/arousal inputs.
%using EmoStyle~\cite{azari2024emostyle}. 
%\seung{In our preliminary experiment, diffusion-based models~\cite{paraperas2024arc2face,ding2023diffusionrig,paskaleva2024unified} were considered; however  we observed that it frequently suffers from preserving the pose and identity of the source images~\cite{huang2024diffusion}, when conditioning on continous emotions.} %we use the StyleGAN2~\cite{karras2020analyzing}-based EmoStyle because %modifying emotions while %diffusion-based models cannot modify emotions with continuous values such as valence and arousal while 
However, when training EmoTalkingGaussian with data obtained from the pre-trained EmoStyle, a mismatch arises between lip movements and speech audio, as EmoStyle does not handle their alignment, as shown in Fig.~\ref{fig:motivate finetune emotstyle}. To resolve this, we propose a lip-aligned emotional face generator to better align the lip movements with the source image while effectively reflecting the intended emotions based on valence/arousal. Furthermore, to mitigate the domain gap between real images and synthetic images generated by our lip-aligned emotional face generator, we apply a loss function that leverages normal maps generated by \cite{Abrevaya_2020_CVPR}. To improve the synchronization of lip movements with in-the-wild audio samples, we use a text-to-speech network~\cite{gTTS} to generate curated speech audio data that is small but diverse in English pronunciation. By incorporating SyncNet~\cite{chung2017out}, we apply a loss function that encourages the alignment between input audio and the image rendered by EmoTalkingGaussian, improving synchronization.

The main contributions are summarized as follows:
\begin{itemize}
    \item We propose EmoTalkingGaussian, an audio-driven talking head generation model that leverages valence and arousal to render continuous emotional expressions without requiring additional data capturing.

    \item We introduce self-supervised learning with a sync loss to improve lip synchronization, utilizing a curated speech audio dataset generated via a text-to-speech network.

    \item Extensive experiments demonstrate that EmoTalkingGaussian effectively renders diverse emotional talking head with lip movements synchronized to the input audio, surpassing the limitations in emotional expressiveness of existing state-of-the-art methods.
\end{itemize}

% Contribution

% Figure 1: Baseline does not manipulate the emotion using the valence and arousal. but ours can do it. Simple EmoStyle do not take account into the lip.

% Small area of valence and arousal in the training video.

% However, training the 3DGS network on a single monocular video is not sufficient to expose the model to a wide range of emotions. This is because the training video typically contains specific emotions tied to particular situations (e.g., speeches, news broadcasts). To address this issue, data containing rich emotional facial expressions while preserving identity should be generated.
\begin{figure*}
    \centering
    \includegraphics[width=0.99\linewidth]{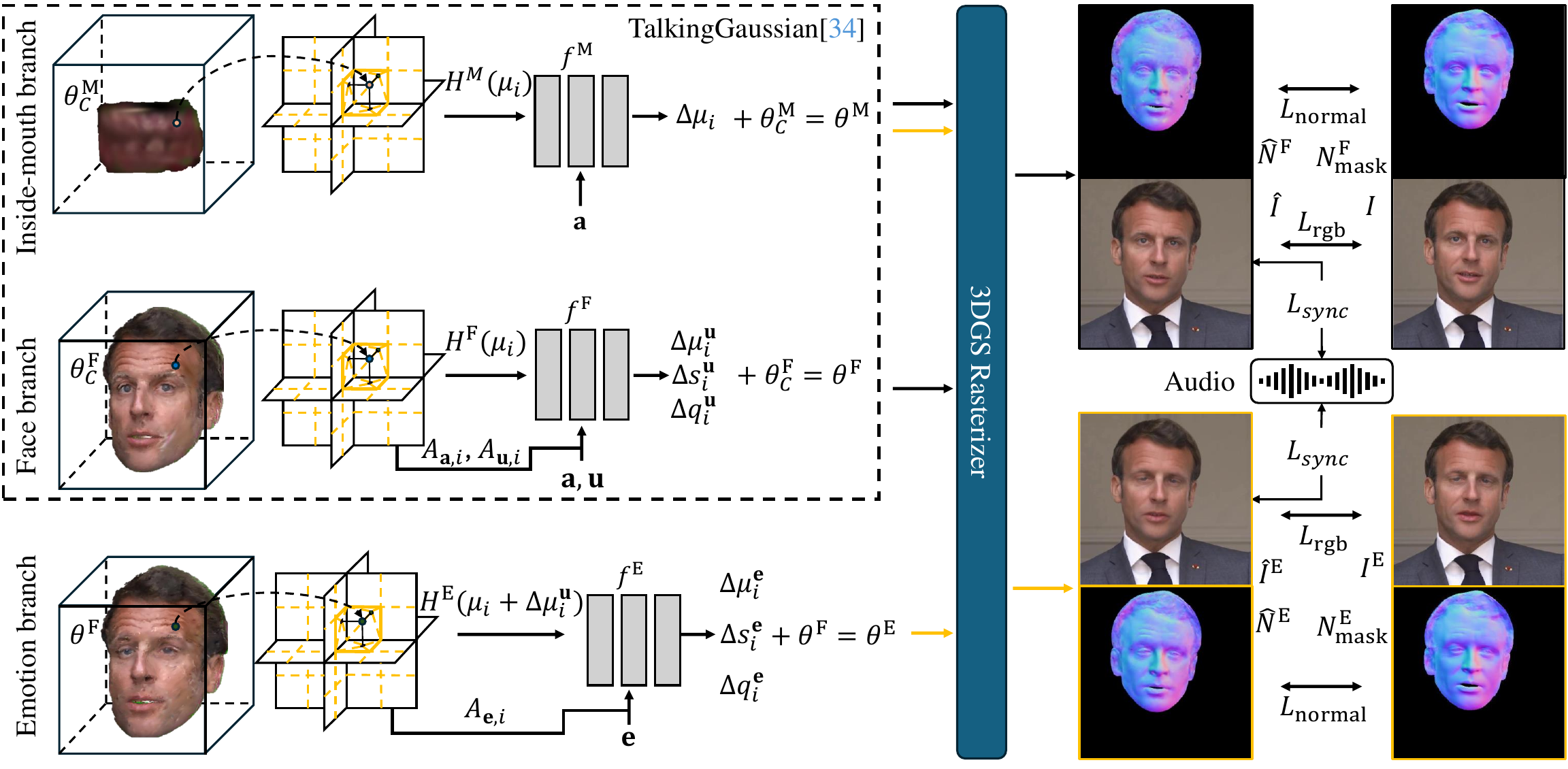}
    \vspace{-3mm}
    \caption{Overview of the EmoTalkingGaussian: Our EmoTalkingGaussian is composed of three branches. First, the inside-mouth branch estimates the position offsets of 3D Gaussians based on audio features $\mba$. Second, the face branch estimates the position, scaling factor, and quaternion offsets based on audio features $\mba$ and action units $\mbu$. Our inside-mouth branch and face branch are inherited from TalkingGaussian~\cite{li2024talkinggaussian}, indicated by the dashed rectangle. Finally, the third branch, the emotion branch, estimates the position, scaling factor, and quaternion offsets based on emotion inputs $\mbe$ (valence/arousal). We render the mouth region and face region $\hat{I}$ along the black arrow. Then, we render the mouth region and emotional face region $\hat{I}^\text{E}$ along the yellow arrow. We apply RGB loss, normal loss, along with audio and lip synchronization loss to improve visual fidelity and overall alignment.}
    \label{fig:pipeline}
    \vspace{-5mm}
\end{figure*}
\section{Related Work}
\label{sec:related_work}
\subsection{3D Gaussian Splatting}
3D Gaussian Splatting (3DGS)~\cite{kerbl20233d} was proposed to address the high computational cost and slow rendering speed issues faced by NeRF~\cite{mildenhall2021nerf}. 3DGS enables fast and efficient rendering, making it highly suitable for real-time applications. 3DGS represents scenes using point-based 3D Gaussians, where each Gaussian contains attributes such as position, scale, rotation, color, and opacity. These Gaussians are aggregated and rendered efficiently using fast differentiable rasterization. In addition to its rendering speed, 3DGS maintains high visual quality, producing detailed and accurate representations of complex scenes, making it suitable for high-fidelity rendering tasks. Thanks to its efficiency and high fidelity, recent research has extended to complex 3D representations of humans~\cite{pang2024ash,jiang2024hifi4g,moon2024exavatar,zhou2024hugs,hu2024gauhuman,pokhariya2024manus,li2024talkinggaussian}, demonstrating potential for diverse applications.

\subsection{Emotional Face Synthesis}
% Emotional face synthesis methods utilize various networks such as Conditional GANs~\cite{mirza2014conditional}, StarGAN~\cite{choi2018stargan}, StyleGAN~\cite{karras2019style}, StyleGAN2~\cite{karras2020analyzing}, and diffusion models~\cite{ho2020denoising,song2020denoising}. 
%
Some emotional face synthesis approaches use Conditional GANs~\cite{mirza2014conditional}, which are conditioned on either a one-hot encoding vector~\cite{ding2018exprgan} or a continuous emotion vector~\cite{lindt2019facial,d2021ganmut}. Lindt~\etal~\cite{lindt2019facial} take valence and arousal values as input to generate corresponding face images. However, they mentioned that preserving the identity becomes difficult if the input image expresses the extreme emotion. Ding~\etal~\cite{d2021ganmut} update valence and arousal vectors so that the generated images are classified as desired emotions.
Some methods~\cite{pumarola2018ganimation,kollias2020va} employ StarGAN~\cite{choi2018stargan} to manipulate the emotion in the input image. Pumarola~\etal~\cite{pumarola2018ganimation} use action units (AUs) to control face muscles, while Kollias~\etal~\cite{kollias2020va} use valence and arousal values to control the emotions of the generated images. StyleGAN~\cite{choi2018stargan} and StyleGAN2~\cite{karras2020analyzing} are utilized in several approaches~\cite{abdal2021styleflow,harkonen2020ganspace,khodadadeh2022latent,patashnik2021styleclip,shen2020interpreting,azari2024emostyle}. 
% Abdal~\etal~\cite{abdal2021styleflow} explored the entangled nature of the GAN latent space to prevent the unwanted distortion by other attributes. 
% Harkonen~\etal~\cite{harkonen2020ganspace} highlighted the importance of the direction in the GAN latent space by applying principal component analysis (PCA). Khodadadeh~\etal~\cite{khodadadeh2022latent} introduced a latent-to-latent mapper and a combination of losses to preserve the identity while editing facial attributes. StyleCLIP~\cite{patashnik2021styleclip} enables facial image editing with text prompts, leveraging CLIP~\cite{radford2021learning}. 
EmoStyle~\cite{azari2024emostyle} uses valence and arousal values to control facial emotions and it proposed a combination of multiple losses related to pixel, landmark, identity, and emotion.
Diffusion~\cite{ho2020denoising,song2020denoising} mechanism-based methods utilize the 3D face mesh model, FLAME~\cite{li2017learning}, to create surface normals, albedo, and Lambertian renderings for conditioning~\cite{ding2023diffusionrig}. These methods~\cite{paskaleva2024unified, paraperas2024arc2face} extend the valence and arousal space into 3D to express more diverse emotions or employ ID vectors to generate desired facial identities.

% Although some existing methods can control emotion using continuous valence and arousal, the mouth region in the generated images often does not align properly with that of the input images. When using these generated images to train 3D Gaussian-based talking head models, this mouth misalignment leads to poor performance in emotional talking head synthesis.

\subsection{Audio-driven Talking Head Synthesis}
2D-based talking head synthesis approaches~\cite{prajwal2020lip,zhang2023dinet,chen2018lip,chung2017out,zhou2019talking,song2018talking,gan2023efficient,wang2023progressive} have advanced by utilizing intermediate representations such as motion~\cite{wang2021audio2head,chen2019hierarchical,wang2022one} and landmarks~\cite{zakharov2019few,zhou2020makelttalk,zhong2023identity,zhong2023identity}. However, these 2D-based methods struggle to maintain naturalness and consistency, especially when there are large changes in head pose.

3D-based talking head synthesis methods~\cite{guo2021ad,li2023efficient,ye2023geneface,shen2022learning,tang2022real,li2024s,cho2024gaussiantalker} utilize neural radiance fields~\cite{mildenhall2021nerf} (NeRF) or 3D Gaussians splatting~\cite{kerbl20233d} (3DGS) to generate a photo-realistic and personalized head models. Recently, TalkingGaussian~\cite{li2024talkinggaussian} proposed the 3DGS-based method for audio-driven talking head synthesis. It employs audio features to synchronize the lips with the input audio and utilizes the action units (AUs) to manipulate facial expressions. However, because it is trained on only a 3-5 minute video, it struggles to represent a continuous and wide range of emotions. This limitation is not exclusive to TalkingGaussian but is a common issue among 3D-based approaches. To address this issue, He~\etal~\cite{he2024emotalk3d} collected EmoTalk3D dataset and proposed \textit{Speech-to-Geometry-to-Appearance} framework. However, because this method is data-driven and person-specific, it has the drawback of requiring data collection for new individuals.
\section{Method}
\label{sec:method}

% Our goal is to control the overall facial expression of a 3D talking head, including emotions, eye blinking, eyebrow movements and lip movements, by conditioning on the continuous valence/arousal values, action units and speech signals. Towards this goal, w

We propose EmoTalkingGaussian, which renders emotional talking heads using 3D Gaussian splatting method, conditioned on valence/arousal values, action units, and audio input: We first employ TalkingGaussian~\cite{li2024talkinggaussian} pipeline to synthesize 3D Gaussian splatting-based talking heads conditioned on audio and action units. Then, we generate lip-aligned emotional face images to effectively train the emotion manipulator of EmoTalkingGaussian with diverse emotional facial images. This simple method leads us to synthesize talking heads reflecting diverse valence/arousal conditions, though the rendering quality slightly diminishes. Furthermore, the lip becomes unsynchronized specifically when conditioned on unseen audio. To relieve the challenges, we improve rendering quality by involving a normal map loss and enhance lip synchronization by employing a sync loss which enforces consistency between the image rendered by EmoTalkingGaussian and the input audio generated by a text-to-speech network~\cite{gTTS}. Our EmoTalkingGaussian framework is shown in Fig.~\ref{fig:pipeline}.

% In this loss, we additionally render normal maps from the 3D Gaussian and enforce its consistency to the normal maps estimated from lip-synchronized emotional images that we generated. Second, we further augment the data to further synchronize lips to unseen speech signals. 

% We involve the text-to-speech (TTS) method~\cite{gTTS} to generate diverse speech signals and input them to our system and enforce the consistency between the input speech signal and the generated images.

%, we decompose the task into two branches: the inside mouth branch and the face branch. The inside mouth branch of the network is conditioned solely on the speech input, ensuring that the inside of the mouth are synchronized with the speech. In contrast, the face branch of the network is conditioned on the emotion values, the speech, and the action units, allowing for precise control of the facial expressions based on the desired emotional state and the speech.

\subsection{3D Gaussian Splatting}
%\noindent \textbf{3D Gaussian Splatting.} 
3D Gaussian splatting~\cite{kerbl20233d} utilizes a set of 3D Gaussians, which are represented by a position $\mu$, a scaling factor $s$, a rotation quaternion $q$, an opacity value $\alpha$, and a color $c$, to describe a 3D scene. In the point-based rendering, at pixel $\mbx_p$, the color $C(\mbx_p)$ and the opacity $\calA(\mbx_p)$ are calculated based on the contributions of a set of 
$N$ Gaussians as follows:
\begin{eqnarray}
    C(\mbx_p) &=& \sum_{i \in N} c_i \tilde{\alpha}_i \prod^{i-1}_{j=1} (1-\tilde{\alpha}_j),\label{eq:gaussian rendering1}\\
    \calA(\mbx_p) &=& \sum_{i \in N} \tilde{\alpha}_i \prod^{i-1}_{j=1} (1-\tilde{\alpha}_j),
    \label{eq:gaussian rendering2}
\end{eqnarray}
where $ \tilde{\alpha}_i = \alpha_i \calG^{proj}_i(\mbx_p), \quad
\calG_i(\mbx) = e^{-\frac{1}{2}(\mbx-\mu_i)^T\Sigma^{-1}_i(\mbx-\mu_i)}$.
% \begin{eqnarray}
%     \Tilde{\alpha}_i = \alpha_i \calG^{proj}_i(\mbx_p), \quad
%     \calG_i(\mbx) = e^{-\frac{1}{2}(\mbx-\mu_i)^T\Sigma^{-1}_i(\mbx-\mu_i)},
%     \label{eq:gaussian rendering3}
% \end{eqnarray}
A covariance matrix $\Sigma_i$ is derived from $s_i$ and $q_i$, and a 2D Gaussian $\calG^{proj}_i$ is the projection of a 3D Gaussian $\calG_i$ onto the image plane. During optimization, 3DGS updates parameters $\theta = \{ \mu, s, q, \alpha, c \}$, and applies both densification and pruning of the 3D Gaussians to find the appropriate number of Gaussians for accurately representing the scene.

Following GaussianShader~\cite{jiang2024gaussianshader}, which proposed a method for rendering normal maps directly from 3D Gaussians, we add a normal residual $\Delta \mbn$ to the 3D Gaussian parameters for the face region, defined as $\theta = \{ \mu, s, q, \alpha, c, \Delta \mbn \}$. This residual refines the normal direction to improve the quality of the rendered normal maps. We then use the rendered normal map to apply the normal map loss.

% Additionally, we render the normal map from the 3D Gaussian for use in the normal consistency loss of EmoTalkingGaussian. GaussianShader~\cite{jiang2024gaussianshader} proposed a method for rendering normal maps directly from 3D Gaussians. In this process, a normal residual $\Delta \mbn$ is required to refine the normal direction, thus it is added to the 3D Gaussian parameters for the face region, defined as $\theta = \{ \mu, s, q, \alpha, c, \Delta \mbn \}$.

\subsection{EmoTalkingGaussian} 
We propose the EmoTalkingGaussian that synthesizes talking heads conditioned on the input audio as well as the continuous emotion and expression values, \ie, valence/arousal and action units. We employ TalkingGaussian~\cite{li2024talkinggaussian} pipeline as our baseline architecture for synthesizing the talking head. It separately models an inside-mouth region and a face region using two distinct persistent Gaussian fields. These fields remain stable and preserve the geometry of the face while allowing dynamic deformations based on input audio features $\mba$ extracted by DeepSpeech~\cite{hannun2014deep} and upper-face action units $\mbu$ extracted by OpenFace~\cite{Baltrusaitis2018openface}. To enable precise control over the deformation of the Gaussians, offsets are calculated using a tri-plane hash encoder $H$~\cite{li2023efficient}, which allows accurate adjustments to the Gaussians' parameters.

For the inside-mouth region, the offset $\delta^m_i=\{ \Delta \mu_i \}$ is estimated via the inside-mouth region manipulation network $f^\text{M}$ conditioned only on $\mba$ as follows:
\begin{eqnarray}
    \delta^m_i = f^\text{M}(H^\text{M}(\mu_i) \oplus \mba), 
    \label{eq:mouth mlp}
\end{eqnarray}
where $\mu_i$ denotes the position of the canonical Gaussian $\theta^\text{M}_C$, and $H^\text{M}(\cdot)$ is the tri-plane hash encoder for the inside-mouth. The inside-mouth deformed Guassians are represented as: $\theta^\text{M}=\theta^\text{M}_C+\delta^m=\{\mu+\Delta \mu, s, q, \alpha, c\}$

For the face region, the offset $\delta^{\mbu}_i=\{\Delta \mu^{\mbu}_i,\Delta s^{\mbu}_i, \Delta q^{\mbu}_i \}$ for each Gaussian is estimated using both $\mba$ and $\mbu$ through the face region manipulation network $f^\text{F}$ as follows:
\begin{eqnarray}
    \delta^{\mbu}_i = f^\text{F}(H^\text{F}(\mu_i) \oplus \mba_{r,i} \oplus \mbu_{r,i}),
    \label{eq:face mlp}
\end{eqnarray}
where $\mba_{r,i}=A_{\mba,i} \odot \mba$ and $\mbu_{r,i}=A_{\mbu,i} \odot \mbu$ represent the region-aware features at position $\mu_i$, and the attention maps $A_{\mba, i}$ and $A_{\mbu, i}$ are derived from $\mba$ and $\mbu$, respectively. $\odot$ and $\oplus$ denote Hadamard product and concatenation, respectively. $\mu_i$ is the position of the canonical Gaussian $\theta^\textbf{F}_C$, and $H^\text{F}(\cdot)$ is the tri-plane hash encoder for the face. The face deformed Gaussians are represented as: $\theta^\text{F}=\theta^\text{F}_C + \delta^{\mbu}=\{\mu+\Delta \mu^{\mbu}, s+\Delta s^{\mbu}, q+\Delta q^{\mbu}, \alpha, c, \Delta \mbn \}$.

% Inspired by Li~\etal~\cite{li2024talkinggaussian}, we decompose the generation branch into the inside mouth branch and the face branch. 
We introduce the emotion branch to manipulate the facial emotion based on continuous valence and arousal values $\mbe$. This allows the emotion manipulation network $f^\text{E}$ to estimate the offset $\delta^{\mbe}_i = \{ \Delta \mu^{\mbe}_i, \Delta s^{\mbe}_i, \Delta q^{\mbe}_i \}$ that aligns with the desired emotion.
% This allows the emotion manipulation network $f^\text{E}$ to generate the appropriate position offset $\Delta \mu^{\mbe}_i$, scaling factor offset $\Delta s^{\mbe}_i$, and rotation quaternion offset $\Delta q^{\mbe}_i$ that align with the desired emotion.
% However, to manipulate the facial emotion, we modify the MLP in face branch so that it can receive not only the audio $\mba$ and action units $\mbe$ but also the valance and arousal values $\mbm$. This allows the network to generate the appropriate position offset $\Delta \mu_i$, scaling factor offset $\Delta s_i$, and rotation quaternion offset $\Delta q_i$ that align with the desired emotion.
\begin{eqnarray}
    \delta^{\mbe}_i = f^\text{E}(H^\text{E}(\mu_i+\Delta \mu^{\mbu}_i) \oplus \mbe_{r,i}),
    \label{eq:emotion mlp}
\end{eqnarray}
where $\mbe_{r,i}=A_{\mbe,i} \odot \mbe$ represents the region-aware features at position $\mu_i+\Delta \mu^{\mbu}_i$ of the deformed Gaussian $\theta^\text{F}$, and $H^\text{E}(\cdot)$ is the tri-plane hash encoder for the emotion. The attention map $A_{\mbe, i}$ is derived from $\mbe$, and $\odot$ and $\oplus$ denote Hadamard product and concatenation, respectively. The emotional deformed Gaussians are expressed as: $\theta^\text{E}=\theta^\text{F} + \delta^{\mbe}=\{\mu+\Delta \mu^{\mbu}+\Delta \mu^{\mbe}, s+\Delta s^{\mbu}+\Delta s^{\mbe}, q+\Delta q^{\mbu}+\Delta q^{\mbe}, \alpha, c, \Delta \mbn \}$.

% The final rendering color combines outputs from both face and in-mouth regions, using Eqs.~\ref{eq:gaussian rendering1}, \ref{eq:gaussian rendering2}, \ref{eq:gaussian rendering3}, and \ref{eq:gaussian rendering4}, as follows:
% \begin{eqnarray}
%     C_{\text{head}}(\mbx_p) &=& C_{\text{face}}(\mbx_p) \times \calA_{\text{face}}(\mbx_p) \nonumber\\
%     &+& C_{\text{mouth}}(\mbx_p) \times (1 - \calA_{\text{face}}(\mbx_p))
%     \label{eq:fused image}
% \end{eqnarray}
% where the face and mouth contributions are blended based on their opacities.

\subsection{Synthetic Image and Audio Augmentation}
When training the EmoTalkingGaussian using the provided personal speech video, the emotion manipulation network $f^\text{E}$ is not able to properly model the emotions for given subjects. Furthermore, the speech audio data is also limited. To relieve the challenges of properly training $f^\text{E}$, the face region manipulation network $f^\text{F}$ and the inside-mouth region manipulation $f^\text{M}$ with rich emotional and speech variations, we involve the synthetic images and audio. Especially, we augment the subject-specific emotional face image by involving a lip-aligned emotional face generator. Also, we use the text-to-speech network~\cite{gTTS} to synthesize new speech audio.
% When training the EmoTalkingGaussian, the face region manipulation network $f^\text{F}$ and the inside-mouth region manipulation network $f^\text{M}$ are mainly trained using the given videos with speech signals. However, the original video signals do not contain much emotional variations and thus, $f^\text{F}$ is not able to properly model the emotions for given subjects. Furthermore, the speech signals are also limited. To relieve the challenges to properly train $f^\text{F}$ and $f^\text{M}$ with rich emotional and speech variations, we involved the synthetic images and audios. Especially, we augment the subject-specific emotional face image by involving lip-algined emotional face generator, as shown in Fig.~\ref{fig:synthetic data}. Also, we use the text-to-speech (TTS) generator~\cite{gTTS} to synthesize new speech signals.

\subsubsection{Lip-aligned Emotional Face Generator} 
\label{sec:lip-aligned emotional face generator}
% Our lip-aligned emotional face generation network $g^\text{LEF}$ needs to generate facial images that reflect both the valence/arousal $\mbe$ as well as lip synchronization to speech signals $\mba$, so that we can further augment the training of $f^\text{E}$ with images displaying diverse emotional expressions and precise lip alignment.
% \textcolor{blue}{which is conditioned on both $\mbe$ and $\mba$.}

To obtain the lip-aligned emotional face generation network $g^\text{LEF}$, we initially adopt the framework of EmoStyle~\cite{azari2024emostyle} that is able to adjust the facial emotions in an input image $I$ conditioned on the valence and arousal $\mbe$, while preserving background, identity and head pose. However, during the adjustment, EmoStyle~\cite{azari2024emostyle} also transforms lips to excessively represent the emotions. 
% This hinders its usage in the talking head synthesis task, since it is essential to ensure the alignment between the lip and the speech. 

To prevent this, we extend EmoStyle~\cite{azari2024emostyle} to be additionally conditioned on the lip landmarks of $I$, ensuring that the lip shape on the generated image closely matches that of $I$. We extract lip heatmaps $H$ from $I$ using the off-the-shelf landmark detector $D_l$~\cite{bulat2017far}. The heatmaps $H$ are then processed by a 2D convolutional encoder $E$, which outputs a lip embedding vector $z_l$. We concatenate the lip embedding vector $z_l$ with an emotional latent code $\calW'$ generated by the original EmoStyle. We then introduce the LipExtract module $M_\text{lip}$ to further process the combined embedding. The LipExtract module outputs a lip modification vector $d_l$, which is added to $\calW'$, resulting in a lip-aligned emotional latent code $\calW''$. This process is expressed as follows:
\begin{eqnarray}
    \calW'' = \calW' + d_l, \ \ d_l = M_\text{lip}(z_l \oplus \calW'), \nonumber \\
    \calW' = EmoStyle(I), \ \ z_l = E(H), \ \ H = D_l(I), 
\end{eqnarray}
where $\oplus$ means concatenation. StyleGAN2~\cite{karras2020analyzing} uses $\calW''$ to generate the synthetic emotional image $I^\text{E}$, aligning the lips with those in $I$ while expressing the desired emotion.

% To prevent this, we extended the framework of EmoStyle~\cite{azari2024emostyle} to be additionally conditioned on the lip landmarks. We extracted heatmaps of the lip $H \in \R^{20 \times 64 \times 64}$ from $I$ using the off-the-shelf landmark detector~\cite{bulat2017far}. These heatmaps are then processed by a 2D convolutional encoder $E$, which outputs a lip embedding vector $e_l \in \R^{64}$. We concatenate $e_l$ with an emotional latent code $\calW' \in \R^{512}$ generated by EmoStyle and we proposed the LipExtract module to further process it. The LipExtract module outputs a lip modification vector $d_l \in \R^{512}$ and it is added to $\calW'$, resulting in a lip-aligned latent code $\calW''$. Then, StyleGANv2~\cite{karras2020analyzing} uses $\calW''$ to generate the final image $\hat{I}$, ensuring that the lip is properly aligned with $I$ while expressing the desired emotion.

We utilize the following loss functions to train the encoder $E$ and the LipExtract module $M_\text{lip}$, and fine-tune the StyleGAN2~\cite{karras2020analyzing}, while freezing other components of Emostyle~\cite{azari2024emostyle}:
\begin{eqnarray}
    \calL = \lambda_1 \cdot \calL_{ll} + \lambda_2 \cdot \calL_{lp} + \lambda_3 \cdot \calL_{reg} + \lambda_4 \cdot \calL_{emo} + \lambda_5 \cdot \calL_{id},
\end{eqnarray}
where lip landmark loss $\calL_{ll}$, lip pixel loss $\calL_{lp}$, and regularization loss $\calL_{reg}$ are defined as follows:
\begin{eqnarray}
    \calL_{ll} = ||\hat{L}_l-L_l||^2_2, \ \ \calL_{lp} = ||\mathbb{M}_l \odot (I^\text{E}-I)||^2_2, \ \ \calL_{reg} = || d_l ||^2_2,
\end{eqnarray}
where $L_l$ and $\hat{L}_l$ represent the lip landmarks estimated from the input image $I$ and the output image $I^\text{E}$ using the landmark detector~\cite{bulat2017far}, respectively. $\mathbb{M}_l$ denotes a rectangle mask created from the lip landmarks $L_l$. The losses $\calL_{ll}$ and $\calL_{lp}$ ensure that the lips in $I$ and $I^\text{E}$ are aligned. The regularization loss $\calL_{reg}$ prevents $d_l$ from diverging. Additionally, the emotion loss $\calL_{emo}$~\cite{azari2024emostyle} guarantees that $I^\text{E}$ reflects the desired emotion, while the identity loss $\calL_{id}$~\cite{azari2024emostyle} ensures that $I^\text{E}$ preserves the identity of $I$. $\lambda_i$ is the weight of each loss term.

\subsubsection{TTS-based Speech Audio Generator}
\label{sec:audio generator}
To enhance the generalizability of lip sync, we employ ChatGPT~\cite{ChatGPT} and a text-to-speech algorithm~\cite{gTTS} to generate curated speech audio data. Specifically, we prompt ChatGPT to create 10 text samples that cover a broad range of English phonetic variations, including essential phonemes and various pronunciation phenomena. For instance, the sentence ``The quick brown fox jumps over the lazy dog" includes most English consonants and vowels, providing comprehensive phoneme coverage. 
% Another example, ``She's going to buy some new clothes at the mall," showcases diphthongs and weakened forms, with "going to" often contracted to "gonna" in casual speech. 
% Additionally, the phrase ``Better late than never, they say" illustrates flapping in American English, where "better" sounds like \textipa{/b\textepsilon \textfishhookr \textschwa r/}.

By converting these texts into speech audio using the text-to-speech network~\cite{gTTS}, we perform self-supervised learning to train EmoTalkingGaussian on a diverse set of speech variations. To do this, we apply a sync loss 
$\calL_\text{sync}$, which calculates the L2 loss between audio features and image features using SyncNet~\cite{chung2017out} as follows:
\begin{eqnarray}
    \calL_\text{sync} = || S_\text{I} (\hat{I}) - S_\text{A} (A) ||^2_2, 
    \label{eq:sync_loss}
\end{eqnarray}
where $S_\text{I}$ and $S_\text{A}$ denote the image encoder and audio encoder of SyncNet. $A$ represents the audio input, and $\hat{I}$ denotes the image rendered by EmoTalkingGaussian, conditioned on $A$. This loss enhances synchronization accuracy, enabling our model to be trained on additional speech audio data without paired RGB video, thus allowing for greater flexibility in data use.

\begin{table*}[t]
    \centering
    \begin{tabular}{l|ccccccc}
    \hline
    Method & PSNR ($\uparrow$) & SSIM ($\uparrow$) & LPIPS ($\downarrow$) & LMD ($\downarrow$) & Sync-E($\downarrow$)/C($\uparrow$) & AUE-U($\downarrow$)/L($\downarrow$) 
    & FPS\\
    \hline
    Ground truth & - & 1 & 0 & 0 & 6.546/7.827 & 0/0 & - \\
    \hline
    % EAT~\cite{}     &  &  &  &  &  &  \\
    % PD-FGC~\cite{}     &  & & &  &  & \\
    % \hline
    % GeneFace~\cite{ye2023geneface} & 31.13 & 0.926 & 0.0561 & 3.120 & 8.732/5.196 & 1.422/1.030 & 23 \\
    ER-NeRF~\cite{li2023efficient} & 33.06 & 0.935 & 0.0274 & 3.110 & 8.443/5.554 & 0.779/0.565 & 34 \\
    GaussianTalker~\cite{cho2024gaussiantalker} & 33.02 & 0.939 & 0.0333 & 3.206 &  8.554/5.741 & 0.766/0.523 & \textbf{121} \\
    TalkingGaussian~\cite{li2024talkinggaussian} & 33.64 & 0.940 & \underline{0.0256} & \underline{2.610} & 8.129/5.919 & 0.279/0.550 & \underline{108} \\
    \hline
    Ours w/o Emo. branch & \textbf{33.87} & \textbf{0.944} & \textbf{0.0255} & \textbf{2.557} & \underline{7.750}/\underline{6.270} & \textbf{0.207}/\textbf{0.515} & 107 \\
    Ours & \underline{33.78} & \underline{0.943} & 0.0267 & 2.638 & \textbf{7.702}/\textbf{6.279} & \underline{0.278}/\underline{0.520} & 101 \\
    \hline
    \end{tabular}
    \caption{We compare quantitative results for self-reconstruction scenario. We highlight the best results in \textbf{bold} and the second-best in \underline{underline}. ``Ours w/o Emo. branch" denotes our method without the emotion branch.}
    \label{tab:scenario 1}
    \vspace{-3mm}
\end{table*}

\subsection{Training}
We independently train each branch of EmoTalkingGaussian (inside-mouth, face, and emotion).
\subsubsection{Optimizing Canonical Gaussians}
We optimize the inside-mouth canonical Gaussians $\theta^\text{M}_C$ and the face canonical Gaussians $\theta^\text{F}_C$ through the L1 loss $L_1$ and D-SSIM loss $L_\text{D-SSIM}$:
\begin{eqnarray}
    \calL_\text{rgb} = \calL_1 (\hat{I}_C, I_\text{mask}) + \gamma_1 \calL_\text{D-SSIM} (\hat{I}_C, I_\text{mask}),
    \label{eq:rgb_loss}
\end{eqnarray}
% \begin{eqnarray}
%     L_\text{rgb} = L_1 (\hat{I}_C, I_\text{mask}) &+& \gamma_1 L_\text{D-SSIM} (\hat{I}_C, I_\text{mask}) \nonumber \\
%     &+& \gamma_2 L_\text{LPIPS}(\hat{I}_C, I_\text{mask}),
%     \label{eq:rgb_loss}
% \end{eqnarray}
where $\hat{I}_C$ represents the image rendered from either $\theta^\text{M}_C$ or $\theta^\text{F}_C$, and $I_\text{mask}$ denotes the masked ground truth for either the inside-mouth region or the face region, where the mask is extracted from the ground truth $I$ following \cite{li2024talkinggaussian}. For optimizing $\theta^\text{F}_C$, the normal map loss is additionally applied to update the positions $\mu$ and the normal residuals $\Delta \mbn$ as follows: 
\begin{eqnarray}
    \calL_\text{normal} = \gamma_2 \calL_1 (\hat{N}^\text{F}_C, N^\text{F}_\text{mask}) + \gamma_3 \calL_\text{tv}(\hat{N}^\text{F}_C) 
    + \gamma_4 || \Delta \mbn ||^2_2,
    \label{eq:normal_loss}
\end{eqnarray}
where $\hat{N}^\text{F}_C$ represents the normal map rendered from 3D Gaussians $\theta^\text{F}_C$, while $N^\text{F}_\text{mask}$ is the masked normal map extracted by the predictor~\cite{Abrevaya_2020_CVPR}. $\calL_\text{tv}$ denotes the total variation loss used to enforce the spatial smoothness in the rendered normal map, and the regularization loss ensures that $\Delta \mbn$ does not diverge. $\gamma_i$ is the weight of each loss term.

\subsubsection{Training Networks for Inside-Mouth and Face Regions}
\label{sec:training network}
We train the tri-plane hash encoder $H$ (inside-mouth $H^\text{M}$ and face $H^\text{F}$) and manipulation network $f$ (inside-mouth $f^\text{M}$ and face $f^\text{F}$) with the following loss function:
\begin{eqnarray}
    \calL = \calL_\text{rgb} + \calL_\text{normal} + \calL_\text{sync},
\end{eqnarray}
where
\begin{eqnarray}
    \calL_\text{rgb} &=& \calL_1 (\hat{I}, I_\text{mask}) + \beta_1 \calL_\text{D-SSIM} (\hat{I}, I_\text{mask}) \nonumber \\
    &+& \beta_2 \calL_\text{LPIPS}(\hat{I}, I_\text{mask}), \\
    \calL_\text{normal} &=& \beta_3 \calL_1 (\hat{N}^\text{F}, N^\text{F}_\text{mask}) + \beta_4 \calL_\text{tv}(\hat{N}^\text{F}) \nonumber\\ 
    &+& \beta_5 || \Delta \mbn ||^2_2,
\end{eqnarray}
where $\hat{I}$ is rendered from either $\theta^\text{M}$ or $\theta^\text{F}$, and $\hat{N}^\text{F}$ is rendered from $\theta^\text{F}$. $\calL_\text{LPIPS}$ denotes the LPIPS loss. Furthermore, we apply a sync loss $\calL_\text{sync}$, defined in Eq.~\ref{eq:sync_loss}, using both original and synthesized audio data to improve lip synchronization accuracy. $\beta_i$ is the weight of each loss term.

% losses $L_\text{rgb}$ and $L_\text{normal}$, defined in Eqs.~\ref{eq:rgb_loss} and \ref{eq:normal_loss}, respectively. This involves replacing $\hat{I}_C$ with $\hat{I}$, which is rendered from $\theta^\text{M}$ or $\theta^\text{F}$, and substituting $\hat{N}^\text{F}_C$ with $\hat{N}^\text{F}$, rendered from $\theta^\text{F}$. Furthermore, we apply a sync loss $L_\text{sync}$, defined in Eq.~\ref{eq:sync_loss}, using both original and synthesized speech data to improve synchronization accuracy.

\subsubsection{Training Network for Emotion}
We train the emotion tri-plane hash encoder $H^\text{E}$ and the emotion manipulation network $f^\text{E}$ with the following loss function:
\begin{eqnarray}
    \calL = \calL_\text{rgb} + \calL_\text{normal} + \calL_\text{sync},
\end{eqnarray}
where
\begin{eqnarray}
    \calL_\text{rgb} &=& \calL_1 (\hat{I}^\text{E}, I_\text{mask}) + \kappa_1 \calL_\text{D-SSIM} (\hat{I}^\text{E}, I_\text{mask}) \nonumber \\
    &+& \kappa_2 \calL_\text{LPIPS}(\hat{I}^\text{E}, I_\text{mask}), \\
    \calL_\text{normal} &=& \kappa_3 \calL_1 (\hat{N}^\text{E}, N^\text{E}_\text{mask}) + \kappa_4 \calL_\text{tv}(\hat{N}^\text{E}) \nonumber\\ 
    &+& \kappa_5 || \Delta \mbn ||^2_2,
\end{eqnarray}
where both $\hat{I}^\text{E}$ and $\hat{N}^\text{E}$ are rendered from $\theta^\text{E}$. Additionally, we apply a sync loss $\calL_\text{sync}$, defined in Eq.~\ref{eq:sync_loss}, using both original and synthesized audio data to enhance the lip synchronization. $\kappa_i$ is the weight of each loss term.

\section{Experiments}
\label{sec:experiments}

\begin{table}[t]
    \centering
    \begin{tabular}{l|c|c}
    \hline
     & Testset A & Testset B \\
    Method & Sync-E($\downarrow$)/C($\uparrow$) & Sync-E($\downarrow$)/C($\uparrow$) \\
    \hline
    % EAT~\cite{}     &  &  &  &   \\
    % PD-FGC~\cite{}     &  & & &   \\
    % \hline
    Ground truth & 7.589/7.158 & 7.398/7.112 \\
    \hline
    % GeneFace~\cite{ye2023geneface} & 9.801/4.410 & 9.274/4.712 \\
    ER-NeRF~\cite{li2023efficient} & 9.960/4.305 & 9.397/4.938 \\
    GaussianTalk.~\cite{cho2024gaussiantalker} & 10.208/4.375 & 9.419/5.001 \\
    TalkingGau.~\cite{li2024talkinggaussian} & 9.369/4.835 & 9.009/5.261 \\
    \hline
    % Ours w/o Emo. & \textbf{9.323}/\textbf{4.964} & \textbf{8.828}/\textbf{5.457}  \\
    Ours & \textbf{9.262}/\textbf{4.930} & \textbf{8.746}/\textbf{5.426}  \\
    \hline
    \end{tabular}
    \caption{We evaluate cross-domain audio scenario, highlighting the best results in \textbf{bold}.}
    \label{tab:scenario 2}
    \vspace{-3mm}
\end{table}

\begin{table*}[t]
    \centering
    \begin{tabular}{l|cccccc}
    \hline
    Method & Sync-E($\downarrow$)/C($\uparrow$) & V-RMSE($\downarrow$) & A-RMSE($\downarrow$) & V-SA($\uparrow$) & A-SA($\uparrow$) & E-Acc($\uparrow$) \\
    \hline
    Ground truth & 7.830/7.042 & - & - & - & - & -\\
    \hline
    % EAT~\cite{}     &  &  &   \\
    % PD-FGC~\cite{}     &  & &   \\
    % \hline
    % GeneFace~\cite{ye2023geneface}  & 10.263/4.208  & 0.494 & 0.489 & 0.500 & 0.500 & 24.2  \\
    ER-NeRF~\cite{li2023efficient}  & 10.109/4.225 & 0.479 & 0.502 & 0.500 & 0.500 & 27.3 \\
    GaussianTalker~\cite{cho2024gaussiantalker}  & 10.243/4.539 & 0.491 & 0.503 & 0.500 & 0.500 & 22.9 \\
    TalkingGaussian~\cite{li2024talkinggaussian}  & 9.580/4.779 & 0.467 & 0.474 & 0.515 & 0.500 & 29.1 \\
    \hline
    % Ours w/o Emo. branch  & \textbf{9.006}/\textbf{5.320} & 0.481 & 0.490 & 0.500 & 0.500 & 25.3 \\
    Ours & \textbf{9.082}/\textbf{5.152} & \textbf{0.352} & \textbf{0.383} & \textbf{0.766} & \textbf{0.637} & \textbf{46.6}  \\
    \hline
    \end{tabular}
    \vspace{-3mm}
    \caption{We compare the models' ability to reflect the desired emotion on the face. The best score is highlighted in \textbf{bold}.}
    \label{tab:scenario 3}
    \vspace{-3mm}
\end{table*}

\subsection{Setup}
\noindent\textbf{Dataset.} We evaluate our method on publicly available videos~\cite{guo2021ad,li2023efficient,ye2023geneface}, following the setting of Li~\etal~\cite{li2024talkinggaussian}. The dataset has $4$ subjects: ``Macron", ``Obama", ``Lieu", and ``May." Each video is cropped and resized to ensure that faces are centered. The average video length is $6,500$ at $25$ FPS, with a resolution of $512\times512$ pixels, except for the ``Obama" video which has $450\times450$ resolution. Each video is divided into train and test sets with a 10:1 ratio.

\noindent\textbf{Baselines.} We compare our method with NeRF-based approaches, ER-NeRF~\cite{li2023efficient}, and 3DGS-based approaches, TalkingGaussian~\cite{li2024talkinggaussian} and GaussianTalker~\cite{cho2024gaussiantalker}. These baseline methods are limited to basic facial expressions, such as eye blinking, and do not enable manipulation of facial emotions. Some 2D-based talking head generation methods~\cite{gan2023efficient,wang2023progressive} require large datasets for training, therefore we do not include them in our comparisons.

\noindent\textbf{Scenarios.} We evaluate methods across three scenarios: \emph{self-reconstruction}, \emph{cross-domain audio}, and \emph{emotion-conditioned} scenarios. In the \emph{self-reconstruction} scenario, we evaluate methods using the test set's audio, action units, and valence/arousal values. In the \emph{cross-domain audio} scenario, we train models on train set and test them on two cross-domain audio samples extracted from \cite{suwajanakorn2017synthesizing}, to evaluate their performance on in-the-wild audio. In the \emph{emotion-conditioned} scenario, we evaluate the models' ability to reflect the desired emotion. For our model, we manipulate emotional expressions by using 12 points selected on a 2D circle as inputs for valence and arousal. For other baseline models, action units are used. Specifically, we generate emotional facial images using EmoStyle~\cite{azari2024emostyle} with the 12 valence and arousal points, then extract action units from these images to use as input. Additionally, we use the cross-domain audio sample extracted from \cite{suwajanakorn2017synthesizing}.
% , which were employed by previous methods~\cite{cho2024gaussiantalker}. 

\noindent\textbf{Metrics.}
We utilize PSNR, SSIM~\cite{wang2004image}, and LPIPS~\cite{zhang2018unreasonable} to evaluate the quality of the rendered images. To evaluate lip synchronization, we use the mouth landmark distance (LMD)~\cite{chen2018lip}, and the synchronization error (Sync-E) and the synchronization confidence score (Sync-C) of SyncNet~\cite{chung2017out}. We measure the upper-face action unit error (AUE-U) and lower-face action unit error (AUE-L) using OpenFace~\cite{baltruvsaitis2015cross,Baltrusaitis2018openface}. These metrics ensure that the rendered images accurately capture and reflect the target facial action units. To evaluate the emotion consistency, we employ the valance and arousal root mean square error (V-RMSE and A-RMSE)~\cite{toisoul2021estimation}, and valance and arousal sign agreement (V-SA and A-SA)~\cite{toisoul2021estimation}. Additionally, we utilize the top-3 emotion classification accuracy (E-Acc). We employ EmoNet~\cite{toisoul2021estimation} to extract valence, arousal, and emotion label from the rendered images.

% The RSME is defined as: $RMSE = \sqrt{\frac{1}{N}\sum^N_i(E^{pred}_i-E^{true}_i)^2}$ where $E^{pred}_i$ and $E^{true}_i$ are the predicted and true emotion values (valence or arousal) for the $i$-th sample. The SA is defined as: $SA=\frac{1}{N}\sum^N_i\mathbb{I}(\text{sign}(E^{pred}_i)==\text{sign}(E^{true}_i))$ where $\mathbb{I}(\cdot)$ is the indicator function that returns 1 if the condition inside is true and 0 otherwise.

\subsection{Quantitative Results}
We evaluate various metrics across three different scenarios, as reported in Tabs~\ref{tab:scenario 1}, \ref{tab:scenario 2}, and \ref{tab:scenario 3}. In \emph{self-reconstruction} scenario, our method without the emotion branch achieves the best scores in pixel-based metrics, such as PSNR, SSIM, and LPIPS, as well as the landmark distance (LMD) and action unit error for the lower face (AUE-L) and upper face (AUE-U). Our method, which incorporates emotion and is trained on both emotional synthetic and original facial data, slightly compromises detail preservation, but it remains comparable to other baselines in pixel-based metrics, and outperforms them with higher lip synchronization confidence and lower action unit error for expressions. In \emph{cross-domain audio} scenario, where we cannot use ground-truth images, we measure synchronization error (Sync-E) and confidence (Sync-C). Our method outperforms other methods in both Sync-E and Sync-C, demonstrating its ability to handle the cross-domain audio effectively. In \emph{emotion-conditioned} scenario, we use Sync-E and Sync-C metrics to evaluate the lip-synchronization, and V-RMSE, A-RMSE, V-SA, and A-SA to evaluate the models' ability to convey the desired emotion to the talking head. Other methods show V-SA and A-SA scores around 0.5, indicating that the valence and arousal values estimated from the rendered images are concentrated in only one quadrant of the valence-arousal circle. Additionally, they exhibit high V-RMSE and A-RMSE errors, along with low emotion classification accuracy. These suggest that these methods do not effectively reflect the desired emotions. In contrast, our model demonstrates superior performance across all metrics, confirming its ability to effectively convey the intended emotions on the face.

\subsection{Qualitative Results}
\begin{figure*}
    \centering
    \includegraphics[width=0.99\linewidth]{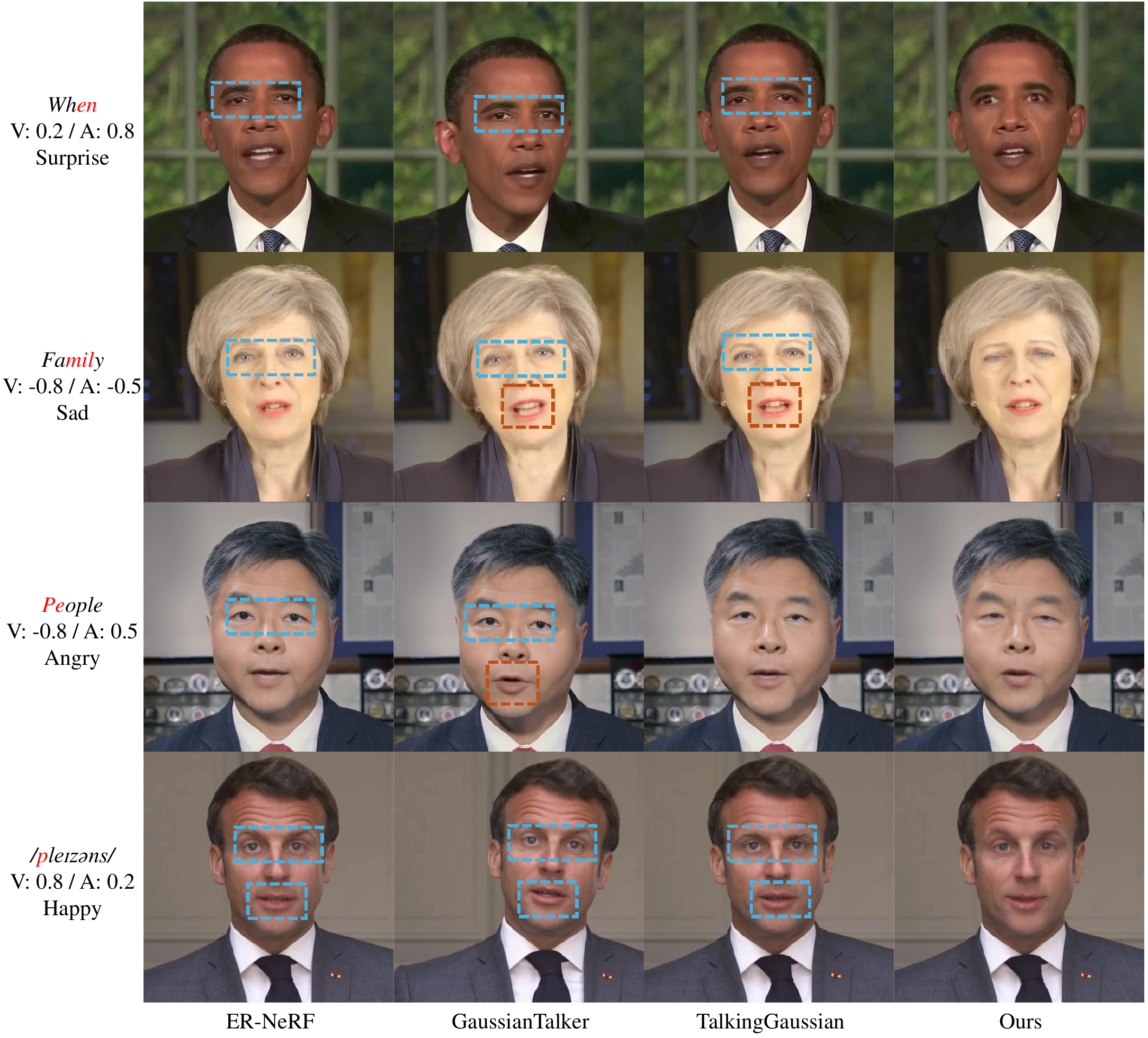}
    \caption{We present qualitative comparisons with other baselines, including ER-NeRF~\cite{li2023efficient}, GaussianTalker~\cite{cho2024gaussiantalker}, and TalkingGaussian~\cite{li2024talkinggaussian}. The word is displayed with the spoken word highlighted in \textcolor{red}{red}. The last sample shows the phonetic transcription. `V' and `A' stand for valence and arousal, and emotion labels indicate the emotion that `V' and `A' values represent. Emotional inconsistencies and lip mismatches are highlighted with \textcolor{skyblue}{blue} and \textcolor{brown}{brown} dashed boxes, respectively.}
    \label{fig:qualitative_results}
    \vspace{-3mm}
\end{figure*}

We present the qualitative results in Fig.~\ref{fig:qualitative_results}. The word pronounced by the subject is highlighted in red. To manipulate facial emotion with our method, we use valence and arousal, shown below the pronounced word as V and A. To better understand which specific emotion the valence and arousal represent, the emotion label is displayed below the V and A values. For other methods, such as ER-NeRF~\cite{li2023efficient}, GaussianTalker~\cite{cho2024gaussiantalker} and TalkingGaussian~\cite{li2024talkinggaussian}, the action units are used to manipulate facial expressions. As shown in the blue dashed boxes in Fig.~\ref{fig:qualitative_results}, action units have a limitation in expressing natural emotions. In contrast, our method expresses emotions around the eyes and mouth based on valence and arousal values. For lip synchronization, mismatches are highlighted with brown dashed boxes. Our method effectively synchronizes lip movements with speech audio while conveying the desired emotions.

\begin{table}[t]
    \centering
    \begin{tabular}{l|cc|c}
    \hline
    Method & LPIPS ($\downarrow$) & Sync-C($\uparrow$) & E-Acc ($\uparrow$) \\
    \hline
    Ours w/o $L_\text{normal}$ & 0.0274 & 6.090 & 45.4 \\
    Ours w/o $L_\text{sync}$ & 0.0269 & 5.925 & 46.1 \\
    \hline
    Combined face b.  & \textbf{0.0253} & 4.489 & 42.1 \\
    Ours w/o emo. b. & 0.0255 & 6.270 & 25.3 \\
    \hline
    Ours & 0.0267 & \textbf{6.279} & \textbf{46.6} \\
    \hline
    \end{tabular}
    \vspace{-3mm}
    \caption{Ablation study on loss functions and model architectures across different model configurations: `Ours w/o $L_\text{normal}$' (without normal map loss), `Ours w/o $L_\text{sync}$' (without sync loss), `Combined face b.' (model combining face branch and emotion branch), `Ours w/o emo. b.' (our model without the emotion branch), and `Ours' (our full model). The best performance for each metric is highlighted in \textbf{bold}.}
    \label{tab:ablation study}
    \vspace{-3mm}
\end{table}

\subsection{Ablation Study}
We conduct ablation study on loss functions and model architectures, as reported in Tab.~\ref{tab:ablation study}. We use the self-reconstruction scenario to measure LPIPS and Sync-C, and the emotion-conditioned scenario to measure E-Acc. For \emph{loss functions}, including the normal map loss $L_\text{normal}$ and the sync loss $L_\text{sync}$, we compare the performance of our full model with `Ours w/o $L_\text{normal}$' and `Ours w/o $L_\text{normal}$'. `Ours w/o $L_\text{normal}$' do not use $L_\text{normal}$ and `Ours w/o $L_\text{sync}$' do not use $L_\text{sync}$, during training of EmoTalkingGaussian. $L_\text{normal}$ impacts the quality of the rendered image, as evidenced by an increase in LPIPS. Similarly, $L_\text{sync}$ influences lip sync accuracy, as reflected in changes in Sync-E/C. Additionally, for \emph{model architectures}, we compare our full model with `Combined face b.' and `Ours w/o Emo. b.'. `Combined face b.' model renders emotional facial images using only the face manipulation network $f^{F}$ in the face branch, where the network $f^{F}$ takes valence, arousal, action units, and audio as inputs. `Ours w/o Emo. b.' represents the our model without the emotion branch. `Combined face b.' achieves the best LPIPS score but struggles with accurate lip synchronization. `Ours w/o Emo. b.' fails to reflect emotion in the rendered image. While our full model shows slightly lower performance in image quality, it outperforms other baseline in both lip synchronization and emotion control.

% \subsubsection{Architectures.}
% We conduct an ablation study on model architectures, including the emotion branch and a new face branch that also incorporates emotion values. 

\section{Conclusion}
\label{sec:conclusion}

This paper introduces a novel 3D emotional talking head generation framework, EmoTalkingGaussian. Our framework can seamlessly utilize even new subject video containing highly sparse emotion representation without any need for additional data capturing. Benefiting from a lip-aligned emotional facial image generator, normal map loss, sync loss, and curated speech audio data, our method enables diverse emotion manipulation based on valence and arousal, synchronizing lip movements in the rendered image with the input audio while preserving high image quality.

\noindent\textbf{Limitation.} Depending on the emotion, the mouth in the synthesized image sometimes change dramatically, causing artifacts around the mouth region in the rendered image by EmoTalkingGuassian. This highlights a trade-off between image fidelity and the intensity of emotional expression.

\noindent\textbf{Ethical consideration.} There is potential for misuse, including in deepfake applications or deceptive media. To mitigate this, we strongly advocate for responsible use, ensuring that generated content is not used for misleading or harmful purposes. We aim to support efforts that aid in the detection and responsible development of deepfake technology.

\setcounter{equation}{0}
\setcounter{figure}{0}
\setcounter{table}{0}
\setcounter{section}{0}
\makeatletter
\renewcommand{\theequation}{S\arabic{equation}}
\renewcommand{\thefigure}{S\arabic{figure}}
\renewcommand{\thetable}{S\arabic{table}}
\renewcommand{\thesection}{S\arabic{section}}
\clearpage
\setcounter{page}{1}
\maketitlesupplementary

In this supplementary material, we provide implementation details; details of lip-aligned emotional face generator; rendering; details of training EmoTalkingGaussian; curated audio data; evaluation in emotion-conditioned scenario; attention visualization; inside-mouth normal map; limitations of simple fused approach; limitations of diffusion model; user study; and qualitative results. Additionally, for animatable results, please refer to the accompanying supplementary video, which includes the emotion-conditioned scenario comparison, valence-arousal interpolation, 360° valence-arousal interpolation (radius: 0.8), and dynamic emotion transitions during speech.

\section{Implementation Details}
Our method is implemented using PyTorch. All experiments are conducted using RTX 4090 GPUs. We train the lip-aligned emotional face generator for 10 epochs using the Adam optimizer with a learning rate of $1\times10^{-4}$. The mouth branch, face branch, and emotion branch are each trained for 50,000 iterations, and the face canonical Gaussians are fine-tuned with an additional 20,000 iterations. We use the AdamW optimizer with a learning rate of $5\times10^{-3}$ for the hash encoder and $5\times10^{-4}$ for the other parts. The learning rates are adjusted using an exponential scheduler. The total training time is 2 hours. The loss weights are described in Secs.~\ref{sec:lef losses} and \ref{sec:ground truth preparation and trainig losses}.
% \begin{eqnarray}
% \{\lambda_1, \lambda_2, \lambda_3, \lambda_4, \lambda_5\} &=& \{ 1, 5, 0.03, 0.2, 1.5 \}, \nonumber \\
% \{\gamma_1, \gamma_2, \gamma_3, \gamma_4\} &=& \{ 0.2, 0.05, 0.005, 0.001 \}, \nonumber \\
% \{\beta_1, \beta_2, \beta_3, \beta_4, \beta_5 \} &=& \{ 0.2, 0.2, 0.05, 0.005, 0.001 \}, \nonumber \\
% \{\kappa_1, \kappa_2, \kappa_3, \kappa_4, \kappa_5 \} &=& \{ 0.2, 0.2, 0.05, 0.005, 0.001 \}. \nonumber
% \end{eqnarray}

\section{Details of Lip-aligned Emotional Face Generator}

\subsection{Pipeline}
\begin{figure}
    \centering
    \includegraphics[width=0.99\linewidth]{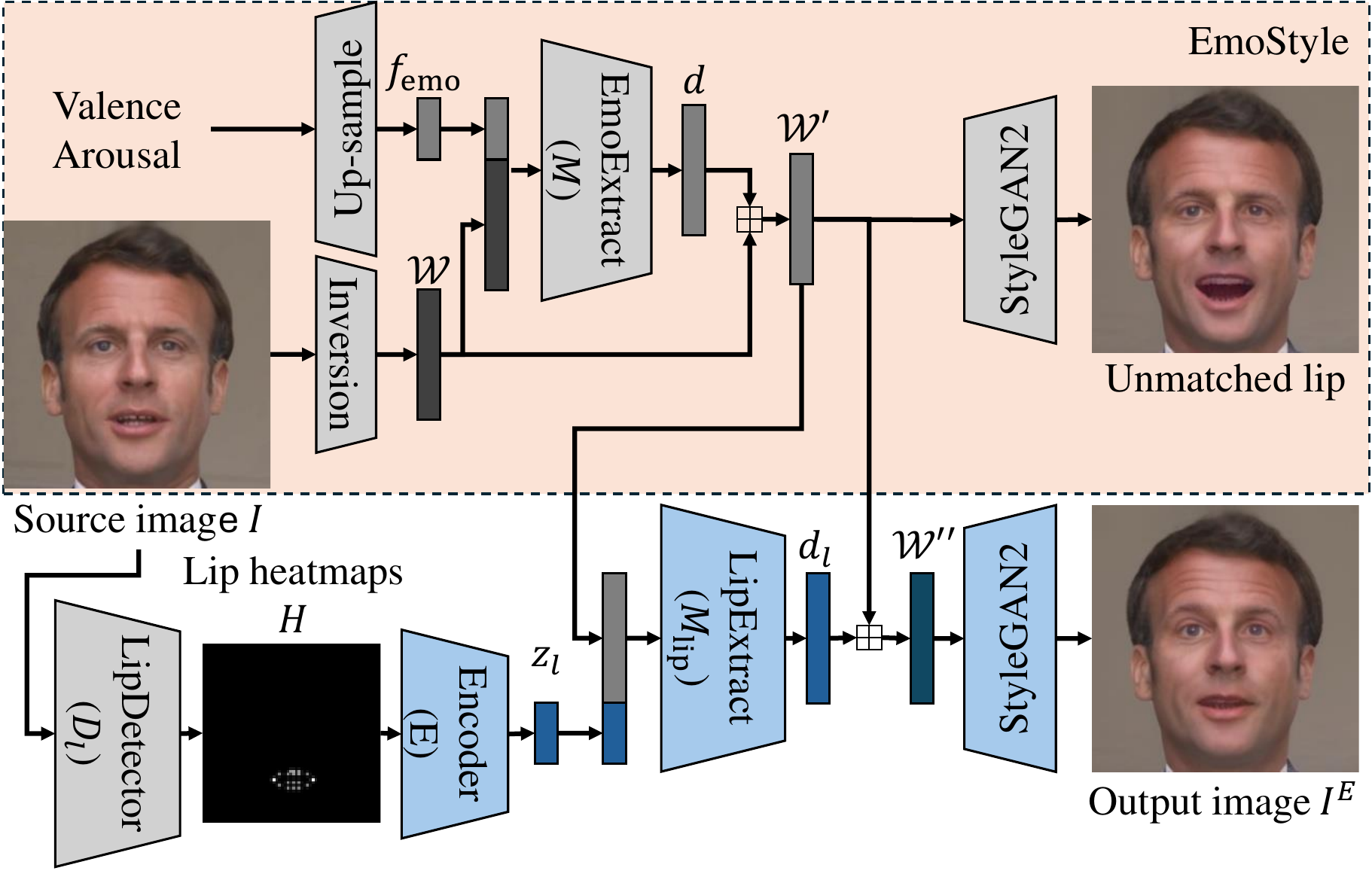}
    \caption{Overview of the lip-aligned emotional face generator. While EmoStyle~\cite{azari2024emostyle} cannot produce lip-aligned emotional facial images, our generator creates such images by aligning lips based on lip heatmaps. $\boxplus$ denotes vector summation.}
    \label{fig:pipeline of face generator}
\end{figure}

The overall framework of our lip-aligned emotional face generator $g^\text{LEF}$ is illustrated in Fig.~\ref{fig:pipeline of face generator}. EmoStyle~\cite{azari2024emostyle} first creates a latent code $\calW$ from a source image $I$ using an inversion module ~\cite{azari2024emostyle,tov2021designing}. It then produces an emotional latent code $\calW'$ by adding $\calW$ to an emotion modification vector $d$. The vector $d$ is generated by EmoExtract $M$, which takes a concatenated vector $(f_\text{emo} \oplus \calW)$ as input. Here, $f_\text{emo}$ represents the valence/arousal features derived from the valence/arousal input $\mbe$ through an up-sampling module. EmoStyle generates a facial image with lips that do not match those of the source image $I$.

To achieve lip alignment, a lip encoder $E$ generates a lip embedding vector $z_l$ from lip heatmaps $H$, which are extracted from the source image $I$ using a LipDetector $D_l$~\cite{bulat2017far}. A LipExtract module $M_\text{lip}$ then produces a lip modification vector $d_l$ by taking a concatenated vector ($z_l \oplus \calW'$) as input. By adding $d_l$ to the emotional latent code $\calW'$, a lip-aligned emotional latent code $\calW''$ is obtained. Finally, our generator employs StyleGAN2~\cite{karras2020analyzing} to generate a lip-aligned emotional facial image $I^\text{E}$ from the latent code $\calW''$.

\begin{figure}
    \centering
    \includegraphics[width=0.99\linewidth]{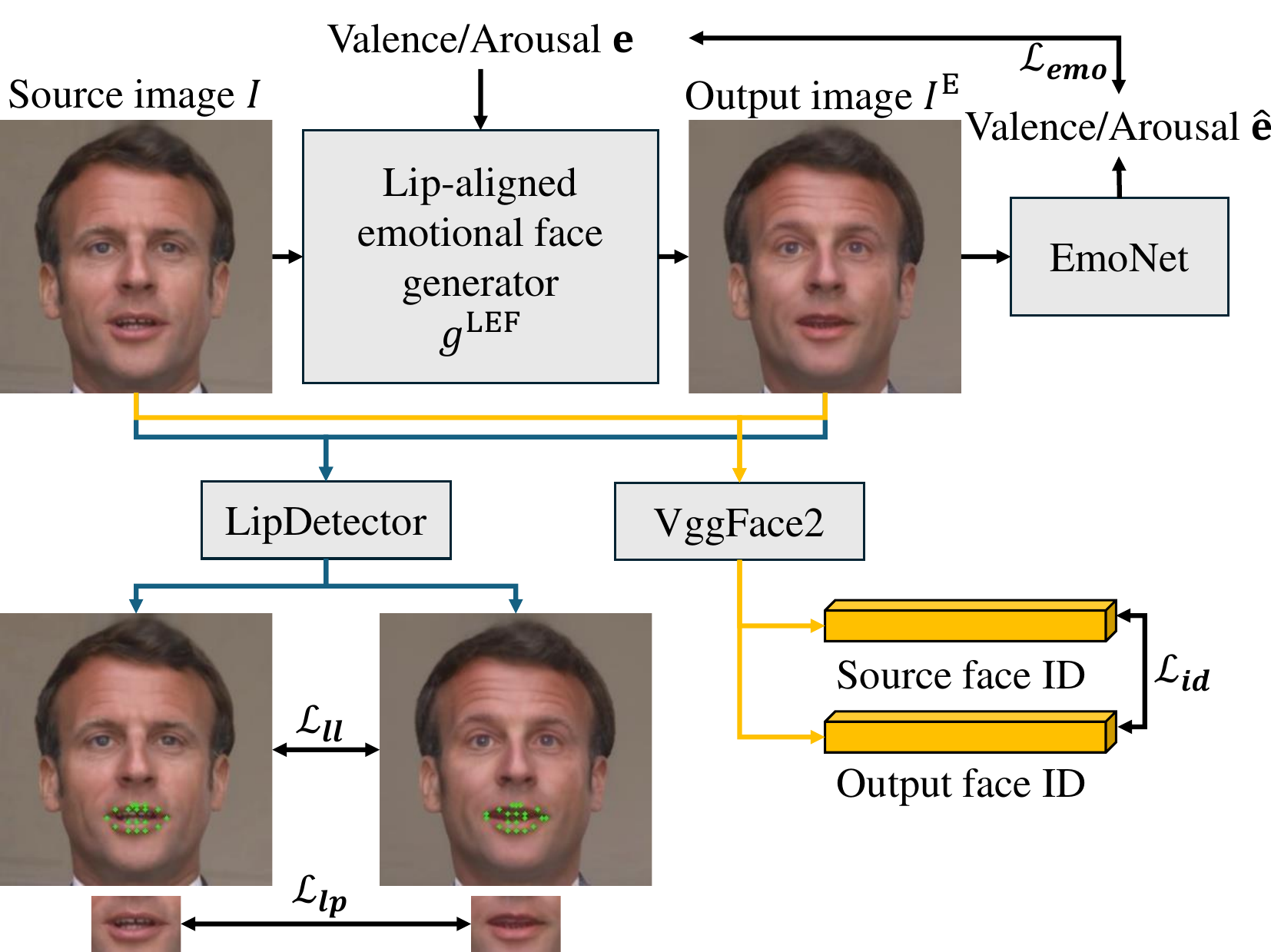}
    \caption{Lip landmark loss $\calL_{ll}$, lip pixel loss $\calL_{lp}$, emotion loss $\calL_{emo}$, identity loss $\calL_{id}$ are utilized to train the lip-aligned emotional face generator $g^\text{LEF}$. We use LipDetector~\cite{bulat2017far}, EmoNet~\cite{toisoul2021estimation}, and VggFace2~\cite{cao2018vggface2}.}
    \label{fig:pipeline of face gen loss}
\end{figure}

\subsection{Losses}
\label{sec:lef losses}
We train the encoder $E$ and the LipExtract module $M_\text{lip}$, and fine-tune StyleGAN2~\cite{karras2020analyzing} using the following loss functions: lip landmark loss, lip pixel loss, regularization loss, emotion loss, and identity loss. These losses are illustrated in Fig.~\ref{fig:pipeline of face gen loss}.

The lip landmark loss $\calL_{ll}$ is defined as follows:
\begin{eqnarray}
    \calL_{ll} = ||\hat{L}_l-L_l||^2_2,
\end{eqnarray}
where $L_l$ and $\hat{L}_l$ denote the lip landmarks extracted from the source image $I$ and the output image $I^\text{E}$, respectively, using the landmark detector~\cite{bulat2017far}. This loss encourages the model to produce an output image $I^\text{E}$ that closely aligns with the original lip structure of $I$, ensuring the positional accuracy of the generated lip region.

The lip pixel loss $\calL_{lp}$ is defined as follows:
\begin{eqnarray}
    \calL_{lp} = ||\mathbb{M}_l \odot (I^\text{E}-I)||^2_2,
\end{eqnarray}
where the lip region mask $\mathbb{M}_l$ is applied to the pixel-wise difference between $I^\text{E}$ and $I$. $\mathbb{M}_l$ is a rectangular mask created based on the lip landmarks $L_l$, and $\odot$ denotes element-wise multiplication. This loss penalizes differences in pixel values within the lip area, encouraging $I^\text{E}$ to closely resemble $I$ specifically in the lip region.

The regularization loss $\calL_{reg}$ is defined as follows:
\begin{eqnarray}
    \calL_{reg} = || d_l ||^2_2.
\end{eqnarray}
This loss prevents $d_l$ from diverging.

The emotion loss $\calL_{emo}$ is defined as follows:
\begin{eqnarray}
    \calL_{emo} = || EmoNet(I^\text{E}) - \mbe ||^2_2,
\end{eqnarray}
where $EmoNet$~\cite{toisoul2021estimation} outputs valence and arousal values $\hat{\mbe}$ from the output image $I^\text{E}$. This loss encourages the output image $I^\text{E}$ to reflect the emotion specified by the input valence and arousal values $\mbe$.

The identity loss $\calL_{id}$ is defined as follows:
\begin{eqnarray}
    \calL_{id} = ||VF(I^\text{E})-VF(I)||_1
\end{eqnarray}
where $VF$ represents VggFace2~\cite{cao2018vggface2}, which extracts identity embeddings corresponding to the face's identity. This loss ensures that the identity of the output image $I^\text{E}$ is preserved, matching that of the source image $I$.

Total training loss are expressed as follows:
\begin{eqnarray}
    \calL = \lambda_1 \cdot \calL_{ll} + \lambda_2 \cdot \calL_{lp} + \lambda_3 \cdot \calL_{reg} \nonumber \\ 
    + \lambda_4 \cdot \calL_{emo} + \lambda_5 \cdot \calL_{id},
\end{eqnarray}
where $\lambda_1$, $\lambda_2$, $\lambda_3$, $\lambda_4$, and $\lambda_5$ are 1, 5, 0.03, 0.2, and 1.5, respectively.

\begin{figure}
    \centering
    \includegraphics[width=0.99\linewidth]{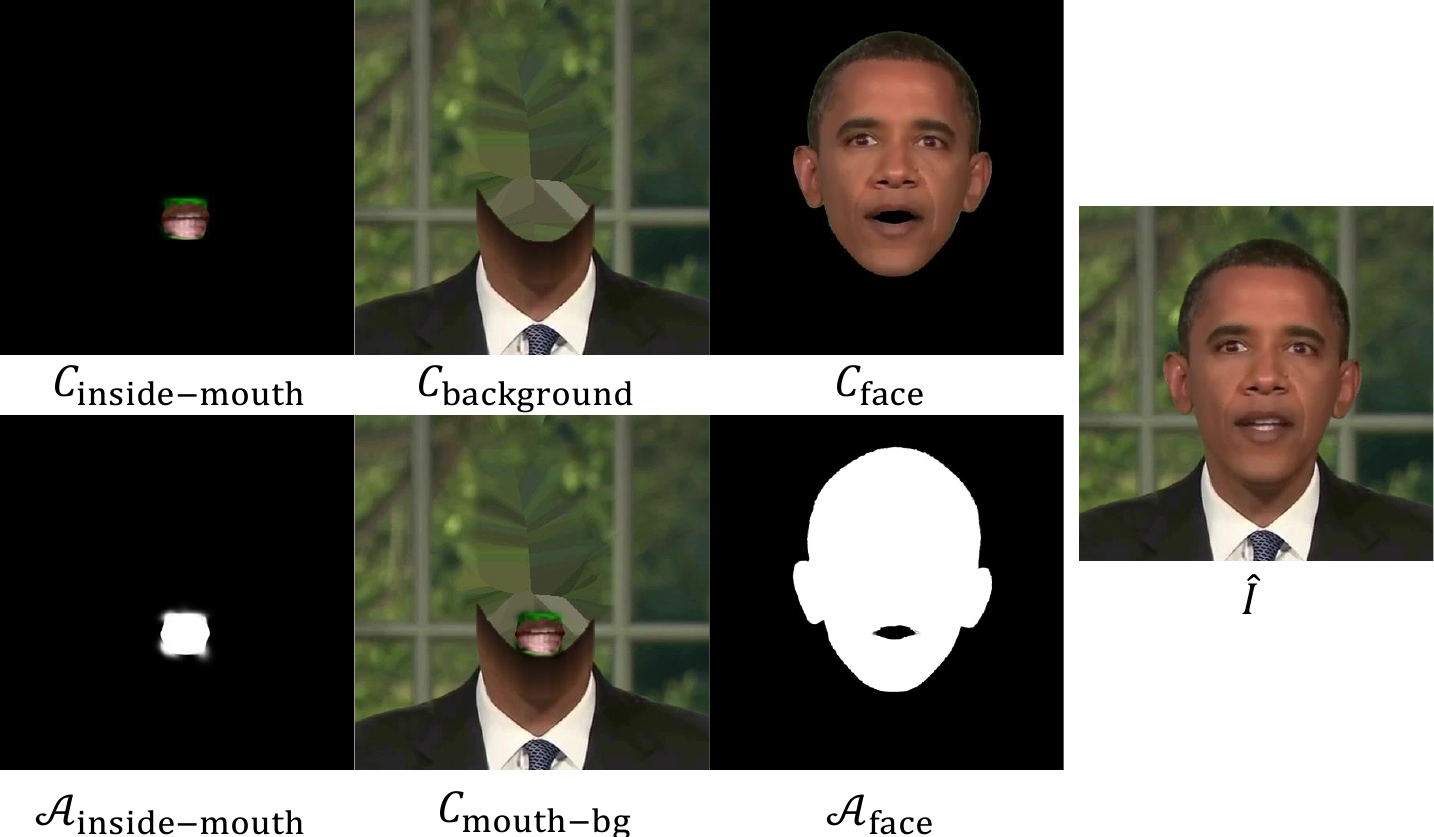}
    \caption{Examples of rendering are displayed, and their relationships are described in Eqs.~\ref{eq:C mouth bg} and \ref{eq:rendering final image}.}
    \label{fig:rendering examples}
\end{figure}

\begin{figure*}
    \centering
    \includegraphics[width=0.9\linewidth]{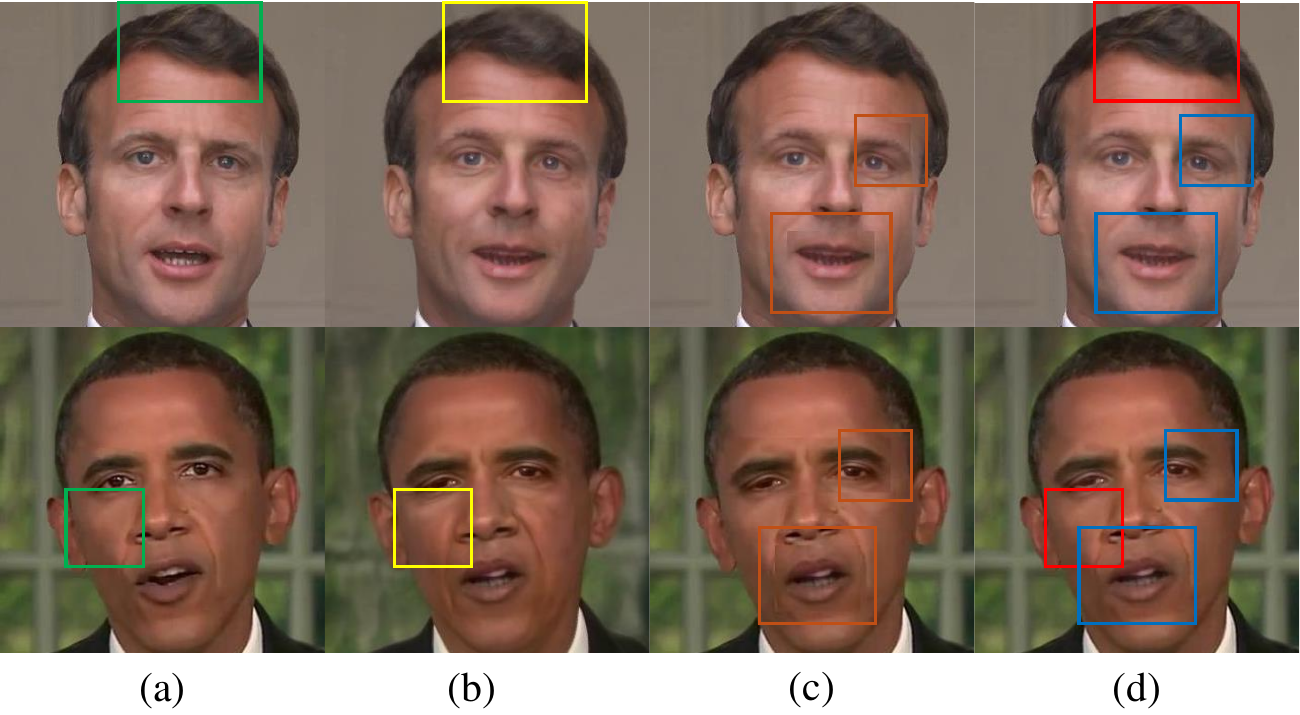}
    \caption{Columns (a), (b), (c), and (d) display source images, images generated by the lip-aligned emotional face generator $g^\text{LEF}$, simple cut-and-paste composite images, and seamless cloning results, respectively. The first row uses valence and arousal values of 0.6 and 0.2 (happy), while the second row applies values of -0.8 and 0.4 (angry). \textcolor{OliveGreen}{Green} and \textcolor{Goldenrod}{yellow} boxes indicate domain gaps, while \textcolor{brown}{brown} boxes highlight boundary artifacts. \textcolor{red}{Red} and \textcolor{blue}{blue} boxes demonstrate that seamless cloning effectively addresses both domain gaps and boundary artifacts, respectively.}
    \label{fig:synthetic data}
\end{figure*}

\section{Rendering}
The $i$-th 3D Gaussian contains a position $\mu_i$, a scaling factor $s_i$, a rotation quaternion $q_i$, an opacity value $\alpha_i$, and a color $c_i$. The rendering process for 3D Gaussians is expressed as follows:
\begin{eqnarray}
    C(\mbx_p) &=& \sum_{i \in N} c_i \tilde{\alpha}_i \prod^{i-1}_{j=1} (1-\tilde{\alpha}_j),\\
    \calA(\mbx_p) &=& \sum_{i \in N} \tilde{\alpha}_i \prod^{i-1}_{j=1} (1-\tilde{\alpha}_j),
\end{eqnarray}
where $C(\mbx_P)$ and $\calA(\mbx_P)$ represent color and opacity at pixel $\mbx_p$, respectively, as described in Eqs. 1 and 2 of the main paper. To generate a final image $\hat{I}$ by combining the face color $C_\text{face}$, inside-mouth color $C_\text{inside-mouth}$, and background color $C_\text{background}$, we use the respective opacities $\calA_\text{face}$ and $\calA_\text{inside-mouth}$ as follows:
\begin{eqnarray}
    C_\text{mouth-bg} &=& C_\text{inside-mouth} \times \calA_\text{inside-mouth} \nonumber \\
    &+& C_\text{background} \times (1-\calA_\text{inside-mouth}),
    \label{eq:C mouth bg}
\end{eqnarray}
and 
\begin{eqnarray}
    \hat{I} &=& C_\text{face} \times \calA_\text{face} + C_\text{mouth-bg} \times (1-\calA_\text{face}),
    \label{eq:rendering final image}
\end{eqnarray}
where we blend the inside-mouth color $C_\text{inside-mouth}$ with the background color $C_\text{background}$ using the opacity $\calA_\text{inside-mouth}$ to produce the combined inside-mouth and background color $C_\text{mouth-bg}$, and then blend this result with the face color $C_\text{face}$, using the opacity $\calA_\text{face}$ to produce the final image $\hat{I}$. The background color $C_\text{background}$ is provided, while $C_\text{inside-mouth}$ and $\calA_\text{inside-mouth}$ are rendered from the inside-mouth deformed Gaussians $\theta^\text{M}$, and $C_\text{face}$ and $\calA_\text{face}$ are rendered from the emotionally deformed Gaussians $\theta^\text{E}$. Examples of rendering are shown in Fig.~\ref{fig:rendering examples}.

\begin{figure*}
    \centering
    \includegraphics[width=0.99\linewidth]{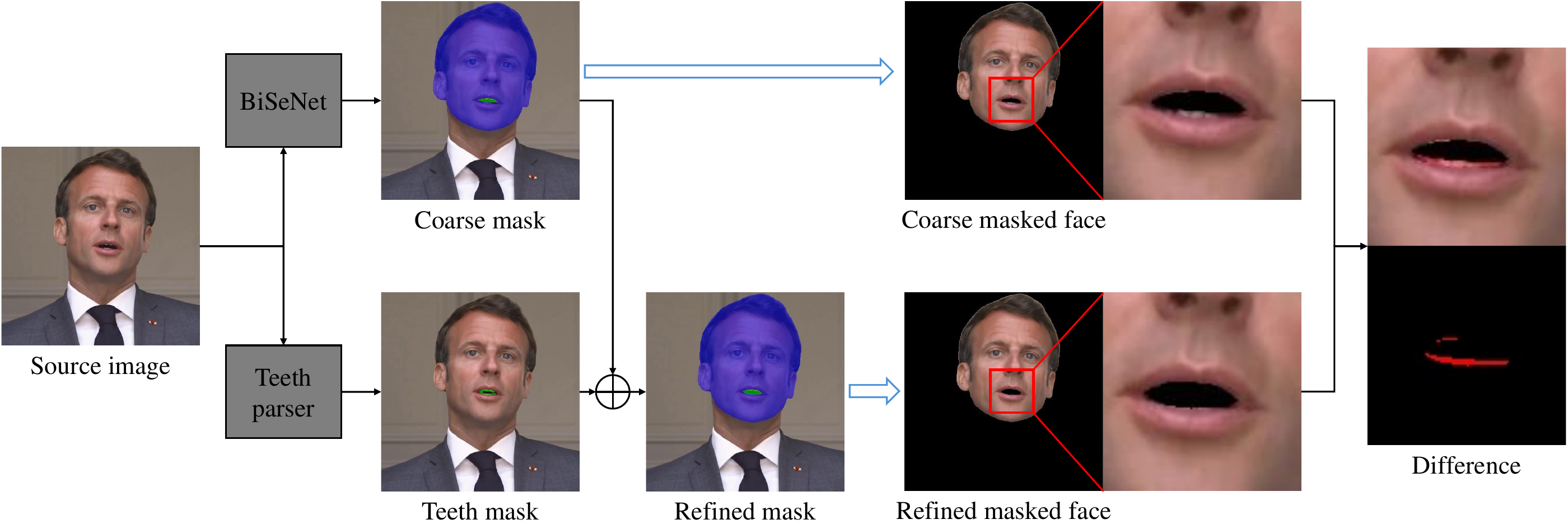}
    \caption{BiSeNet~\cite{yu2018bisenet} estimates the coarse mask to parse the face and mouth regions, while the Teeth parser~\cite{li2024talkinggaussian} estimates the teeth mask. These two masks are combined to create the refined mask, which more accurately separates the face and mouth compared to the coarse mask. The areas of discrepancy between the coarse and refined masks are highlighted in red, denoted as ``Difference", representing pixel differences between the face regions masked by each individual mask.}
    \label{fig:processing mask}
\end{figure*}

\section{Details of Training EmoTalkingGaussian}
\subsection{Improve Synthetic Emotional Facial Image}
\label{sec:seamless cloning}
The initial generated image is produced by our lip-aligned emotional face generation network $g^\text{LEF}$, manipulates a source image by taking any valence/arousal $\mbe$ as input to generate an emotional face image that align with $\mbe$. However, the generated images exhibit domain gaps compared to real images, particularly noticeable in artifacts such as blurry hair or irregularities in skin texture, as seen in the green boxes in column (a) and the yellow boxes in column (b) of Fig.~\ref{fig:synthetic data}. To reduce this gap, we use a cut-and-paste composite method. Rectangular regions around the eyes and mouth, which are expressive of emotion in the generated image, are cut and pasted onto the source image, as shown in column (c) of Fig.~\ref{fig:synthetic data}. These rectangular regions are determined using eye and mouth landmarks extracted from the generated image. However, while this approach reduces the domain gap by preserving realism in the source image and transferring emotion-expressive regions, it introduces boundary artifacts around the pasted areas, as shown in the brown boxes in column (c) of Fig.~\ref{fig:synthetic data}. To resolve this, we apply seamless cloning~\cite{perez2023poisson} that allows for smooth blending between the two images. Unlike the simple cut-and-paste composite, which results in visible boundary artifacts, seamless cloning integrates textures and colors more naturally, effectively eliminating hard edges and creating a cohesive, realistic appearance, as seen in the red and blue boxes in column (d) of Fig~\ref{fig:synthetic data}. This realistic synthesis is leveraged for the training of our EmoTalkingGaussian.

\subsection{Processing Mask}
\label{sec:processing mask}
Following TalkingGaussian~\cite{li2024talkinggaussian}, we utilize BiSeNet~\cite{yu2018bisenet} and Teeth parser~\cite{li2024talkinggaussian} to extract face and inside-mouth masks from a source image, as shown in Fig.~\ref{fig:processing mask}. BiSeNet, pre-trained on CelebAMask-HQ dataset~\cite{lee2020maskgan}, generates a coarse mask which distinguishes face and mouth regions. By masking with the blue region of the coarse mask on the source image, as indicated by the above blue arrow, we extract the coarse masked face. However, as shown, this coarse mask cannot accurately separate the face and mouth regions, especially the teeth. Thus, to address this issue, the Teeth parser, trained on the EasyPortrait dataset~\cite{kvanchiani2023easyportrait}, is employed to generate a teeth mask. By combining the coarse mask and the teeth mask, we create a refined mask. When the source image is masked using the blue region of the refined mask, as indicated by the below blue arrow, the teeth are no longer visible. The mask shown in red highlights the differences between the coarse masked face and the refined masked face. The green region of the refined mask is used to extract the inside-mouth region in Sec.~\ref{sec:ground truth preparation and trainig losses}.

\subsection{Ground Truth Preparation and Details of Training Losses}
\label{sec:ground truth preparation and trainig losses}
\begin{figure*}[t]
    \centering
    \includegraphics[width=0.99\linewidth]{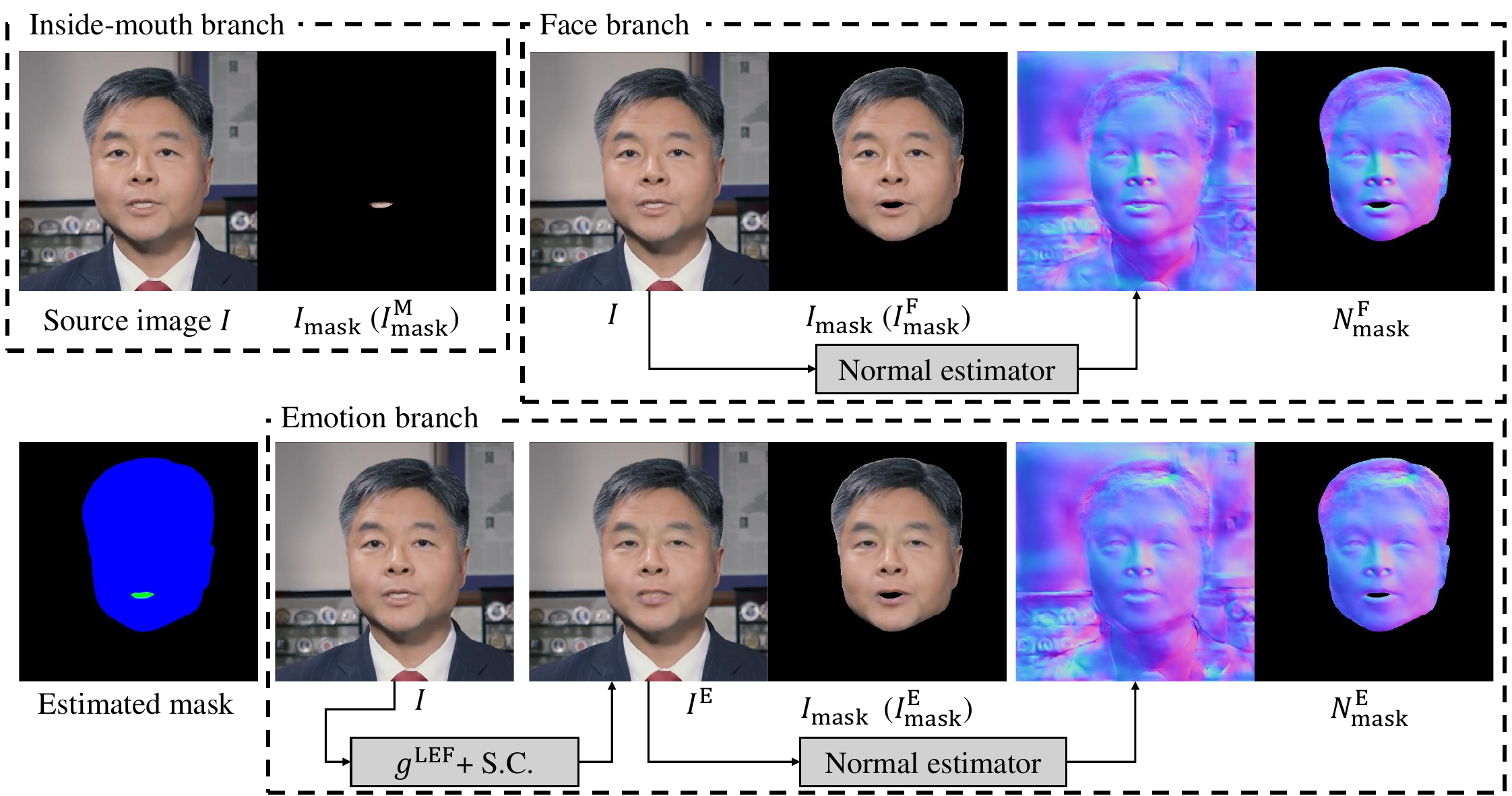}
    \caption{Preparation of ground truth data for training EmoTalkingGaussian involves the lip-aligned emotional face generator $g^\text{LEF}$ combined with seamless cloning (described in Sec.~\ref{sec:seamless cloning} and `S.C.' stands for seamless cloning), the normal estimator~\cite{Abrevaya_2020_CVPR}, and the estimated mask (described in Sec.~\ref{sec:processing mask}).}
    \label{fig:processing data}
\end{figure*}

We prepare the ground truth data for each branch: the inside-mouth branch, face branch, and emotion branch, as shown in Fig.~\ref{fig:processing data}. 

For the inside-mouth branch, we extract an inside-mouth region $I_\text{mask} (I^\text{M}_\text{mask})$ from a source image $I$ using a green region of the mask estimated in Sec.~\ref{sec:processing mask}. The inside-mouth RGB image $I^\text{M}_\text{mask}$ is used as $I_\text{mask}$ in Eqs.~11 and 14 of the main paper. Accordingly, the inside-mouth canonical Gaussians $\theta^\text{M}_C$ are optimized using the following loss function:
\begin{eqnarray}
    \calL = \calL_1 (\hat{I}^\text{M}_C, I^\text{M}_\text{mask}) + \gamma_1 \calL_\text{D-SSIM} (\hat{I}^\text{M}_C, I^\text{M}_\text{mask}),
\end{eqnarray}
where $\hat{I}^\text{M}_C$ denotes the RGB image rendered from $\theta^\text{M}_C$, and $\gamma_1$ is 0.2. The inside-mouth region tri-plane hash encoder $H^\text{M}$ and the inside-mouth region manipulate network $f^\text{M}$ are trained using the following loss function:
\begin{eqnarray}
    \calL &=& \calL_1 (\hat{I}^\text{M}, I^\text{M}_\text{mask}) + \beta_1 \calL_\text{D-SSIM} (\hat{I}^\text{M}, I^\text{M}_\text{mask}) \nonumber \\
    &+& \beta_2 \calL_\text{LPIPS} (\hat{I}^\text{M}, I^\text{M}_\text{mask}) + \beta_6 \calL_\text{sync},
\end{eqnarray}
where $\hat{I}^\text{M}$ denotes the RGB image rendered from inside-mouth deformed Gaussians $\theta^\text{M}$, and $\beta_1$, $\beta_2$, and $\beta_6$ are 0.2, 0.2, and 0.05, respectively.

For the face branch, we extract a face region $I_\text{mask} (I^\text{F}_\text{mask})$ from the source image $I$ using the blue region of the estimated mask. Additionally, we create a masked face normal map $N^\text{F}_\text{mask}$ using a normal estimator~\cite{Abrevaya_2020_CVPR} along with the estimated mask. The face RGB image $I^\text{F}_\text{mask}$ is used as $I_\text{mask}$ in Eqs.~11 and 14 of the main paper, while the normal map $N^\text{F}_\text{mask}$ is utilized in Eqs.~12 and 15 of the main paper. Accordingly, the face canonical Gaussians $\theta^\text{F}_C$ are optimized using the following loss function:
\begin{eqnarray}
    \calL &=& \calL_1 (\hat{I}^\text{F}_C, I^\text{F}_\text{mask}) + \gamma_1 \calL_\text{D-SSIM} (\hat{I}^\text{F}_C, I^\text{F}_\text{mask}) \nonumber\\
    &+& \gamma_2 \calL_1 (\hat{N}^\text{F}_C, N^\text{F}_\text{mask}) + \gamma_3 \calL_\text{tv} (\hat{N}^\text{F}_C) \nonumber \\
    &+& \gamma_4 || \Delta n ||,
\end{eqnarray}
where $\hat{I}^\text{F}_C$ and $\hat{N}^\text{F}_C$ denote the RGB image and normal map rendered from $\theta^\text{F}_C$, and $\gamma_1$, $\gamma_2$, $\gamma_3$, and $\gamma_4$ are 0.2, 0.05, 0.005, and 0.001, respectively. The face region tri-plane hash encoder $H^\text{F}$ and the face region manipulate network $f^\text{F}$ are trained using the following loss function:
\begin{eqnarray}
    \calL &=& \calL_1 (\hat{I}^\text{F}, I^\text{F}_\text{mask}) + \beta_1 \calL_\text{D-SSIM} (\hat{I}^\text{F}, I^\text{F}_\text{mask}) \nonumber \\
    &+& \beta_2 \calL_\text{LPIPS} (\hat{I}^\text{F}, I^\text{F}_\text{mask}) + \beta_3 \calL_1 (\hat{N}^\text{F}, N^\text{F}_\text{mask}) \nonumber \\
    &+& \beta_4 \calL_\text{tv} (\hat{N}^\text{F}) + \beta_5 || \Delta n || + \beta_6 \calL_\text{sync},
\end{eqnarray}
where $\hat{I}^\text{F}$ and $\hat{N}^\text{F}$ denote the RGB image and normal map rendered from face deformed Gaussians $\theta^\text{F}$, and $\beta_1$, $\beta_2$, $\beta_3$, $\beta_4$, $\beta_5$, and $\beta_6$ are 0.2, 0.2, 0.05, 0.005, 0.001, and 0.05, respectively.

For the emotion branch, we generate the emotional facial image $I^\text{E}$ using our lip-aligned emotional face generator $g^\text{LEF}$ and seamless cloning algorithm described in Sec.~\ref{sec:seamless cloning}. Similarly to the previous branches, we extract the RGB $I_\text{mask} (I^\text{E}_\text{mask})$ and normal map $N^\text{E}_\text{mask}$ for the face region using the estimated mask and the normal estimator~\cite{Abrevaya_2020_CVPR}. The emotional face RGB image $I^\text{E}_\text{mask}$ is used as $I_\text{mask}$ in Eq.~17 of the main paper, while the emotional face normal map $N^\text{E}_\text{mask}$ is utilized in Eq.~18 of the main paper. Accordingly, the emotion tri-plane hash encoder $H^\text{E}$ and the emotion manipulation network $f^\text{E}$ are trained using the following loss function:
\begin{eqnarray}
    \calL &=& \calL_1 (\hat{I}^\text{E}, I^\text{E}_\text{mask}) + \beta_1 \calL_\text{D-SSIM} (\hat{I}^\text{E}, I^\text{E}_\text{mask}) \nonumber \\
    &+& \beta_2 \calL_\text{LPIPS} (\hat{I}^\text{E}, I^\text{E}_\text{mask}) + \beta_3 \calL_1 (\hat{N}^\text{E}, N^\text{E}_\text{mask}) \nonumber \\
    &+& \beta_4 \calL_\text{tv} (\hat{N}^\text{E}) + \beta_5 || \Delta n || + \beta_6 \calL_\text{sync},
\end{eqnarray}
where $\hat{I}^\text{E}$ and $\hat{N}^\text{E}$ denote the RGB image and normal map rendered from emotionally deformed Gaussians $\theta^\text{E}$, and $\beta_1$, $\beta_2$, $\beta_3$, $\beta_4$, $\beta_5$, and $\beta_6$ are 0.2, 0.2, 0.05, 0.005, 0.001, and 0.001, respectively.

After finishing training, the face canonical Gaussians $\theta^\text{F}_C$ are further optimized to erase artifacts around face border using the following loss function, focusing only on optimizing the Gaussian's opacity $\alpha$ and color $c$:
\begin{eqnarray}
    \calL = \calL_1 (\hat{I}, I) &+& \eta_1 \calL_\text{D-SSIM} (\hat{I}, I) \nonumber \\
    &+& \eta_2 \calL_\text{LPIPS} (\hat{I}, I),
\end{eqnarray}
where $\hat{I}$ represents the image rendered by Eq.~\ref{eq:rendering final image}, $\eta_1$ and $\eta_2$ are 0.2 and 0.5, respectively.

\section{Curated Audio Data}
We synthesize curated audio data to improve lip synchronization by using ChatGPT~\cite{ChatGPT} and text-to-speech network~\cite{gTTS}. The 10 text descriptions of the audio are listed as follows:
% \begin{enumerate}
%     \item \textbf{Sentence 1}: ``The quick brown fox jumps over the lazy dog."
%     \item \textbf{Sentence 2}: ``She's going to buy some new clothes at the mall."
%     \item \textbf{Sentence 3}: ``I can't believe it's already half past eight!"
%     \item \textbf{Sentence 4}: ``Do you want to grab some water?"
%     \item \textbf{Sentence 5}: ``Better late than never, they say."
%     \item \textbf{Sentence 6}: ``She needs to see the doctor immediately."
%     \item \textbf{Sentence 7}: ``Did you eat yet?"
%     \item \textbf{Sentence 8}: ``Next stop is Central Park."
%     \item \textbf{Sentence 9}: ``An unknown number called me yesterday."
%     \item \textbf{Sentence 10}: ``We can meet at the café if you'd like."
% \end{enumerate}

\begin{description}
    \item \textbf{Sentence 1.} The quick brown fox jumps over the lazy dog.
    \begin{itemize}
        \item \textbf{Phonetic Notation:} \textipa{/Di:/, /kwIk/, /braUn/, /fA:ks/, /dZ2mps/, /"oUv\textrhookschwa/, /Di:/, /"leIzi/, /dAg/}.
        \item \textbf{Sound Coverage:} Includes nearly all English consonants and vowels.
        \item \textbf{Liaison:} /r/ liaison occurs in ``over the" (\textipa{/oUv@ D@/}).
        \item \textbf{Stress:} Content words such as ``quick", ``brown", ``fox", ``jumps", ``over", ``lazy", and ``dog" are stressed.
        \item \textbf{Weak Form:} ``The" (\textipa{/D@/}).
    \end{itemize}

    \item \textbf{Sentence 2.} She's going to buy some new clothes at the mall.
    \begin{itemize}
        \item \textbf{Phonetic Notation:} \textipa{/Si:z/, /"goUIN/, /tu:/, /baI/, /s2m/, /nu:/, /kloUDz/, /\ae t/, /Di:/, /mA:l/}.
        \item \textbf{Diphthongs:} \textipa{/aI/} (buy), \textipa{/oU/} (clothes).
        \item \textbf{Weak Forms:} 
            \begin{itemize}
                \item ``to": \textipa{/t@/}.
                \item ``some": \textipa{/s@m/}.
                \item ``at": \textipa{/@t/}.
                \item ``the": \textipa{/D@/}.
            \end{itemize}
    \end{itemize}

    \item \textbf{Sentence 3.} I can't believe it's already half past eight!
    \begin{itemize}
        \item \textbf{Phonetic Notation:} \textipa{/aI/, /k\ae nt/, /bI"li:v/, /Its/, /A:l"redi/, /h\ae f/, /p\ae st/, /eIt/}.
        \item \textbf{Contractions:} ``can't" as \textipa{/k\ae nt/}, ``it's" as \textipa{/Its/}.
        \item \textbf{Stress:} ``can't", ``believe", ``already", ``half", and ``eight" are stressed.
        \item \textbf{Silent Letters:} The ``l" in ``half" is silent.
    \end{itemize}

    \item \textbf{Sentence 4.} Do you want to grab some water?
    \begin{itemize}
        \item \textbf{Phonetic Notation:} \textipa{/du:/, /ju:/, /wA:nt/, /tu:/, /gr\ae b/, /s2m/, /wA:\v*t\textrhookschwa/}.
        \item \textbf{R-pronunciation:} ``water" is pronounced as \textipa{/"wA:\v*t\textrhookschwa/}.
        \item \textbf{Weak Forms:}
            \begin{itemize}
                \item ``to": \textipa{/t@/}.
                \item ``some": \textipa{/s@m/}.
            \end{itemize}
    \end{itemize}

    \item \textbf{Sentence 5.} Better late than never, they say.
    \begin{itemize}
        \item \textbf{Phonetic Notation:} \textipa{/be\v*t\textrhookschwa/, /leIt/, /D \ae n/, /"nev\textrhookschwa/, /DeI/, /seI/}
        \item \textbf{Flapping:} ``better" sounds like \textipa{/"be\v*t\textrhookschwa/} with a flapped t.
        \item \textbf{Liaison:} Connection in ``than never" with n.
        \item \textbf{Weak Form:} ``than" (\textipa{D@n}).
    \end{itemize}

    \item \textbf{Sentence 6.} She needs to see the doctor immediately.
    \begin{itemize}
        \item \textbf{Phonetic Notation:} \textipa{/Si:/, /ni:ds/, /tu:/, /si:/, /Di:/, /"dA:kt\textrhookschwa/, /I"mi:di@tli/}
        \item \textbf{Long Vowel:} Long \textipa{/i:/} sounds in ``needs" and ``see".
        \item \textbf{Stress:} ``needs", ``see", ``doctor", and ``immediately" are stressed.
        \item \textbf{Weak Forms:}
            \begin{itemize}
                \item ``to": \textipa{/t@/}.
                \item ``the": \textipa{/D@/}.
            \end{itemize}
    \end{itemize}

    \item \textbf{Sentence 7.} Did you eat yet?
    \begin{itemize}
        \item \textbf{Phonetic Notation:} \textipa{/dId/, /ju:/, /i:t/, /jet/}
        \item \textbf{Liaison /j/:} ``Did you" sounds like \textipa{/dIdZu:/}.
        \item \textbf{Stress:} ``eat" is stressed.
    \end{itemize}

    \item \textbf{Sentence 8.} Next stop is Central Park.
    \begin{itemize}
        \item \textbf{Phonetic Notation:}  \textipa{/nekst/, /stA:p/, /Iz/, /"sentr@l/, /pA:rk/}
        \item \textbf{Consonant Cluster:} \textipa{/"sentr@l/} in ``central".
        \item \textbf{Elision:} The ``t" is dropped in ``next stop", sounding like \textipa{/nEks stAp/}.
        \item \textbf{Stress:} ``next", ``stop", ``Central", and ``Park" are stressed.
    \end{itemize}

    \item \textbf{Sentence 9.} An unknown number called me yesterday.
    \begin{itemize}
        \item \textbf{Phonetic Notation:} \textipa{/\ae n/, /2n"n@Un/, /"n2mb\textrhookschwa/, /kA:ld/, /mi:/, /"jest\textrhookschwa deI/}
        \item \textbf{Double Consonant:} The n is lengthened in ``unknown" (\textipa{/n"n/}).
        \item \textbf{R-pronunciation:} The word ``number" has an r-colored schwa vowel (\textipa{\textrhookschwa r}).
        % \item \textbf{Syllable Insertion:} A schwa \textrhookschwa is inserted between \textipa{/b/} and \textipa{/r/} in ``number".
        \item \textbf{Stress:} ``unknown", ``number", ``called", and ``yesterday" are stressed.
        \item \textbf{Weak Form:} ``An" (\textipa{@n}).
    \end{itemize}

    \item \textbf{Sentence 10.} We can meet at the café if you'd like.
    \begin{itemize}
        \item \textbf{Phonetic Notation:} \textipa{/wi:/, /k \ae n/, /mi:t/, /\ae t/, /Di:/, /k\ae f"eI/, /If/, /ju:d/, /laIk/}
        \item \textbf{Contraction:} ``you'd" pronounced as \textipa{/ju:d/}.
        \item \textbf{Stress:} ``meet", ``café", and ``like" are stressed.
        \item \textbf{Weak Forms:}
            \begin{itemize}
            \item ``can": \textipa{/k@n/}.
            \item ``at": \textipa{/@t/}.
            \item ``the": \textipa{/D@/}.
            \end{itemize}
    \end{itemize}
\end{description}

\section{Evaluation in Emotion-conditioned Scenario}
We evaluate our method against state-of-the-art approaches across various scenarios. While prior studies~\cite{li2023efficient,cho2024gaussiantalker,li2024talkinggaussian} primarily focused on self-reconstruction and cross-domain audio scenarios, this paper introduces the emotion-conditioned scenario for the first time. This new scenario allows us to evaluate each method's ability to accurately reflect the desired emotion in the rendered face. To evaluate them, we first select 12 valence and arousal values:

\noindent\([0.74, 0.31], [0.31, 0.74], [-0.31, 0.74], [-0.74, 0.31],\)
\\
\([-0.74, -0.31], [-0.31, -0.74], [0.31, -0.74],\) \\
\([0.74, -0.31], [0.35, 0.35], [-0.35, 0.35],\) \\
\([-0.35, -0.35], [0.35, -0.35]\).
Specifically, the valence-arousal points with a radius of 0.8 are selected by dividing 360° into 8, with each point separated by a 45° interval. Similarly, the valence-arousal points with a radius of 0.5 are chosen by dividing 360° into 4, placing each point at a 90° interval. This approach ensures that the valence-arousal points are evenly distributed around the circle.

Using these valence-arousal points, our model conveys emotion directly. In contrast, other methods~\cite{li2023efficient,cho2024gaussiantalker,li2024talkinggaussian} cannot utilize valence-arousal values directly. Instead, they rely on EmoStyle~\cite{azari2024emostyle} to generate each individual's emotional face based on valence-arousal points. Action units are then extracted from this generated image using OpenFace~\cite{Baltrusaitis2018openface}, which are subsequently used to adjust facial expressions when rendering the talking heads in these methods~\cite{li2023efficient,cho2024gaussiantalker,li2024talkinggaussian}.

\begin{figure}[t]
    \centering
    \includegraphics[width=0.99\linewidth]{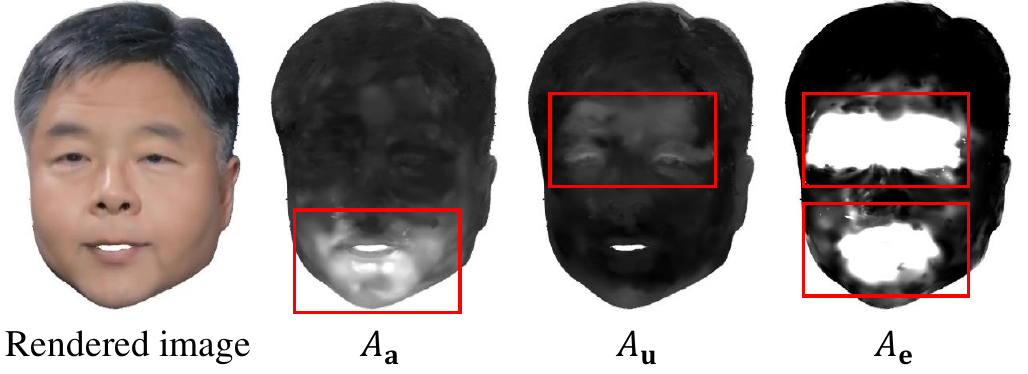}
    \caption{Attention visualization. $A_\mba$, $A_\mbu$, and $A_\mbe$ represent attention maps for audio, action units, and valence/arousal, respectively.}
    \label{fig:attn vis}
\end{figure}

We measure V-RMSE, A-RMSE, V-SA, and A-SA. The root mean square error RMSE is defined as: 
\begin{eqnarray}
    \text{RMSE} = \sqrt{\frac{1}{N}\sum^N_{i=1}(\calE^{pred}_i-\calE^{true}_i)^2},
\end{eqnarray}
where $\calE^{true}_i$ denotes the selected valence or arousal point used as the condition, $\calE^{pred}_i$ represents the valence or arousal value estimated by EmoNet~\cite{toisoul2021estimation} for the $i$-th frame sample, and $N$ denotes the number of frames.

The sign agreement SA is defined as: 
\begin{eqnarray}
    \text{SA} = \frac{1}{N}\sum^N_{i=1}\mathbb{I}(\text{sign}(\calE^{pred}_i)==\text{sign}(\calE^{true}_i)),
\end{eqnarray}
where $\mathbb{I}(\cdot)$ is the indicator function that returns 1 if the condition inside is true and 0 if false. The $\text{sign}(\cdot)$ function outputs 1 if the input is positive, -1 if it is negative, and 0 if it is zero.

Additionally, we utilize frame-wise emotion classification accuracy. Since assigning a single precise emotion class label to each valence-arousal pair is challenging, we evaluate the performance using top-3 accuracy. EmoNet~\cite{toisoul2021estimation} predicts emotion class labels from images rendered by each talking head synthesis model, and the accuracy is measured by comparing the predicted class labels with the predefined emotion class labels. Predefined emotion class labels are assigned to each valence-arousal point as follows:

\noindent\([0.74, 0.31]: \ \text{Happy}, \quad [0.31, 0.74]: \ \text{Surprise},\) \\
\([-0.31, 0.74]: \ \text{Angry}, \quad [-0.74, 0.31]: \ \text{Disgust},\) \\
\([-0.74, -0.31]: \ \text{Sad}, \quad [-0.31, -0.74]: \ \text{Sad}, \) \\
\([0.31, -0.74]: \ \text{Contempt}, \quad [0.74, -0.31]: \ \text{Contempt}, \) \\
\([0.35, 0.35]: \ \text{Happy}, \quad [-0.35, 0.35]: \ \text{Angry}, \) \\
\([-0.35, -0.35]: \ \text{Sad}, \quad [0.35, -0.35]: \ \text{Contempt}\).

\section{Attention Visualization}
We apply an attention mechanism before feeding the audio features $\mba$ and action units $\mbu$ into the manipulation network $f^\text{F}$, as described in Eq.~5 of the main paper, and the valence/arousal $\mbe$ into the emotion manipulation network $f^\text{E}$, as described in Eq.~6 of the main paper. The attention maps, denoted as $A_\mba$, $A_\mbu$, and $A_\mbe$, are visualized in Fig.~\ref{fig:attn vis}. The audio attention map $A_\mba$ focuses primarily on the regions around the lips and jaw, capturing movements conditioned on input speech audio. The action units attention map $A_\mbu$ emphasizes areas around the eyelids and eyebrows, highlighting expressive changes based on action units, such as eye blinking and eyebrow movements. Meanwhile, the emotion attention map $A_\mbe$ concentrates on the mouth and eyes, emphasizing features crucial for reflecting emotional expressions.

\begin{figure}
    \centering
    \includegraphics[width=0.99\linewidth]{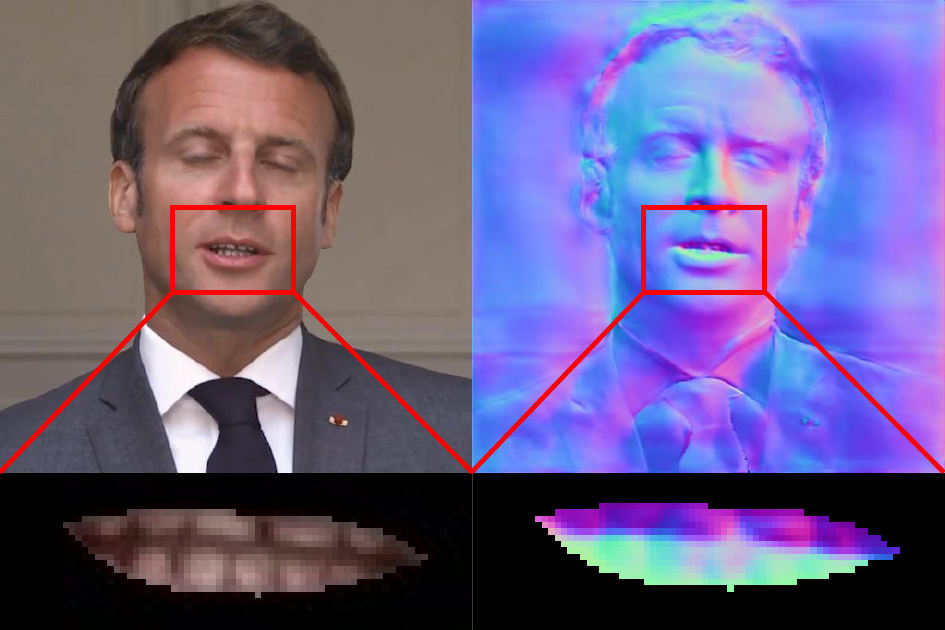}
    \caption{The normal map is estimated by the method in \cite{Abrevaya_2020_CVPR}. The row below shows zoomed-in RGB and normal map images of the inside-mouth region, masked with the refined mask described in Sec.~\ref{sec:processing mask}.}
    \label{fig:inside-mouth normal map}
\end{figure}

\section{Inside-mouth Normal Map}
Fig.~\ref{fig:inside-mouth normal map} presents an RGB image and a normal map of the inside-mouth region. The normal map is estimated by \cite{Abrevaya_2020_CVPR}. The normal map of the inside-mouth region is not well estimated, as shown by the inconsistency between the upper and lower teeth. It leads to unstable optimization of the 3D canonical Gaussians $\theta^\text{M}_C$, along with unstable training of the tri-plane hash encoder $H^\text{M}$ and the manipulation network $f^\text{M}$ for inside-mouth region.

\section{Limitations of Simple Fused Approach}
\begin{figure}
    \centering
    \includegraphics[width=0.99\linewidth]{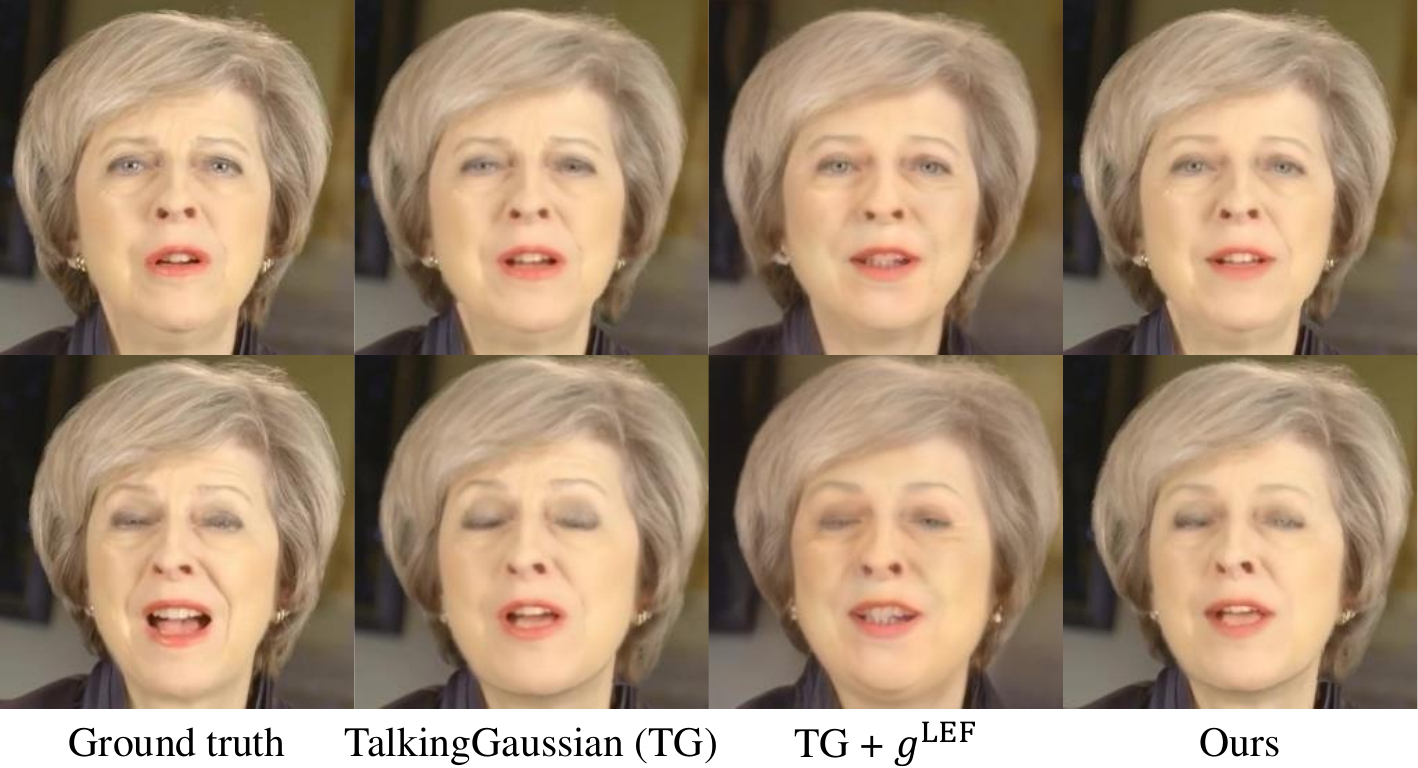}
    \caption{The results of directly applying our lip-aligned emotional face generator $g^\text{LEF}$ to the results of TalkingGaussian (TG)~\cite{li2024talkinggaussian}.}
    \label{fig:direct emostyle}
\end{figure}

We fuse TalkingGaussian~\cite{li2024talkinggaussian} with our lip-aligned emotional face generator $g^\text{LEF}$ and visualize the results in Fig.~\ref{fig:direct emostyle}. Although our face generator $g^\text{LEF}$ adjusts emotions based on a valence/arousal setting of (0.5, 0.1), which corresponds to a slight smile, it does not align the teeth shape with the output of TalkingGaussian.

\section{Limitations of Diffusion-based Model}
\begin{figure}[t]
    \centering
    \includegraphics[width=0.99\linewidth]{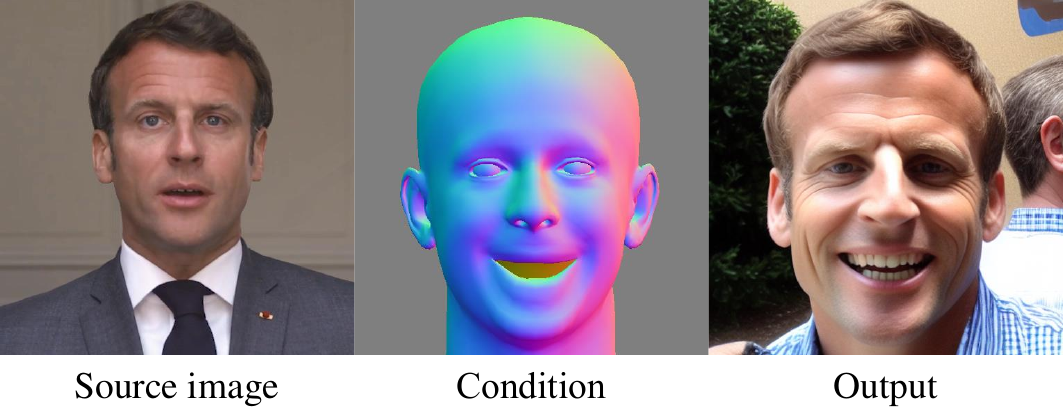}
    \caption{Result from diffusion-based models, Arc2Face~\cite{paraperas2024arc2face}, and ControlNet~\cite{zhang2023adding}.}
    \label{fig:diffusion}
\end{figure}

There are two primary approaches to generate emotional facial images: GAN-based models and diffusion-based models. In our task, there are numerous possible combinations of valence, arousal, and video frames, making it impractical to generate and store emotional facial images in advance. Therefore, we generate the images during the training of EmoTalkingGaussian. However, diffusion-based face generation models are inherently slow, which inevitably increases training time. Additionally, these models tend to significantly alter the pose and style of the source face image, often resulting in a lack of synchronization with the source image and producing unrealistic appearances. To modify the emotion in facial images using diffusion-based models, we use Arc2Face~\cite{paraperas2024arc2face} and ControlNet~\cite{zhang2023adding}, which require conditions rendered from the FLAME model~\cite{li2017learning}. We utilize the normal map rendered from the FLAME mesh as the condition. The FLAME mesh incorporates pose and shape parameters obtained from the source image, along with expression parameters that represent smiling-related features. Although these methods enable changes in emotion, as shown in Fig.~\ref{fig:diffusion}, the resulting face undergoes significant alterations and appears unnatural. Therefore, we utilize GAN-based models for more efficient and effective generation of emotional facial images.

\section{User Study}
In Table~\ref{tab:user study}, we report the results of a user study conducted to evaluate the reflection of emotions, lip synchronization to the audio, and overall quality. While lip synchronization performance is comparable across models, our method demonstrates the highest capability in reflecting the desired emotions and receives the most selections for overall quality.

\begin{table}[h]
    \centering
    \begin{tabular}{l|ccc}
        \hline
          & SD1 (\%) & SD2 (\%) & SD3 (\%) \\
         \hline
         ER-NeRF~\cite{li2023efficient} & 4.17 & 15.83 & 16.25 \\
         GaussianTalker~\cite{cho2024gaussiantalker} & 10.00 & 31.67 & 25.84 \\
         TalkingGaussian~\cite{li2024talkinggaussian} & 5.42 & 14.58 & 11.25 \\
         Ours & \textbf{69.58} & \textbf{32.92} & \textbf{43.33} \\
         none & 10.83 & 5.00 & 3.33 \\
         \hline
    \end{tabular}
    \caption{The proportion of videos selected by users for each evaluation criterion is presented in this table. `SD1' denotes Standard 1, representing the proportion of videos chosen by users as best reflecting emotions. `SD2' represents Standard 2, indicating the proportion of videos selected as best in lip synchronization to the audio, and `SD3' denotes Standard 3, reflecting the proportion of videos rated as best in overall quality.}
    \label{tab:user study}
\end{table}

\section{Qualitative Results}
\begin{figure*}[t]
    \centering
    \includegraphics[width=0.99\linewidth]{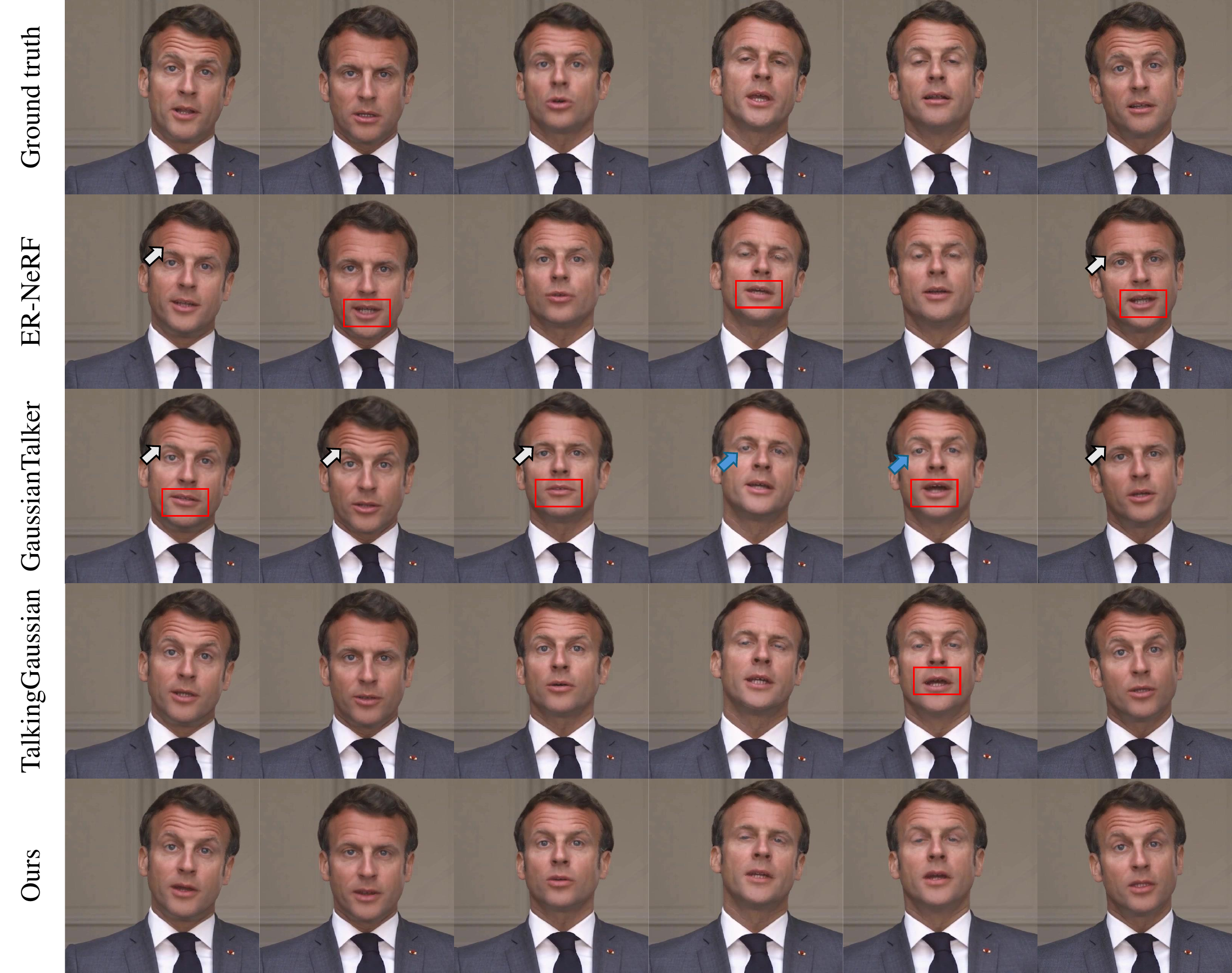}
    \caption{We present the qualitative comparisons in the self-reconstruction scenario against other methods, including ER-NeRF~\cite{li2023efficient}, GaussianTalker~\cite{cho2024gaussiantalker}, and TalkingGaussian~\cite{li2024talkinggaussian}. Misalignment with the ground truth is highlighted using \textcolor{gray}{gray} arrows for discrepancies in the eyebrows and forehead wrinkles, \textcolor{blue}{blue} arrows for blinking errors, and \textcolor{red}{red} boxes for lip misalignment.}
    \label{fig:scenario1}
\end{figure*}

\begin{figure*}[t]
    \centering
    \includegraphics[width=0.99\linewidth]{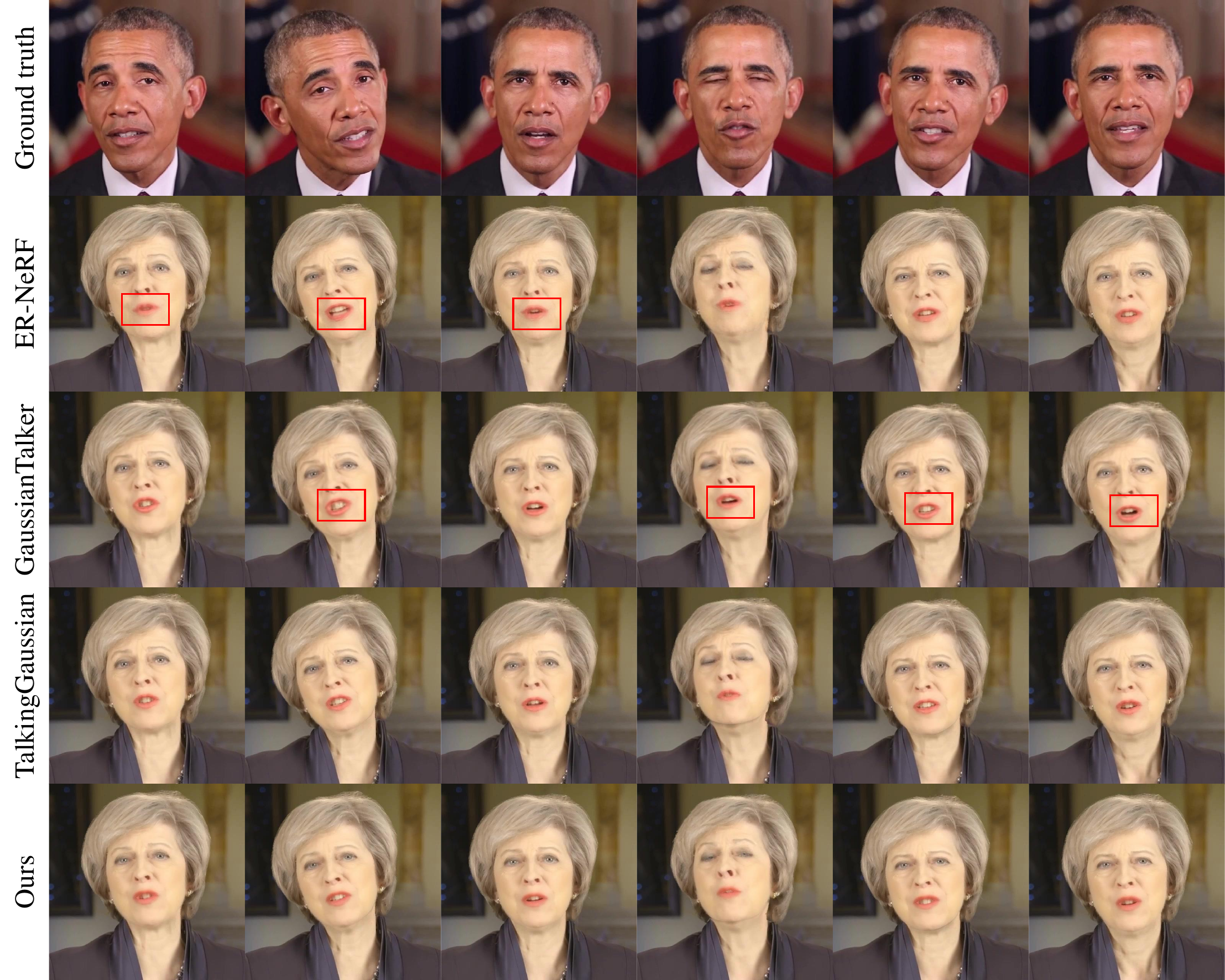}
    \caption{We present qualitative comparisons in the cross-domain audio scenario against other methods, including ER-NeRF~\cite{li2023efficient}, GaussianTalker~\cite{cho2024gaussiantalker}, and TalkingGaussian~\cite{li2024talkinggaussian}. Lip misalignment with the ground truth is highlighted using \textcolor{red}{red} boxes.}
    \label{fig:scenario2}
\end{figure*}

\begin{figure*}[t]
    \centering
    \includegraphics[width=0.99\linewidth]{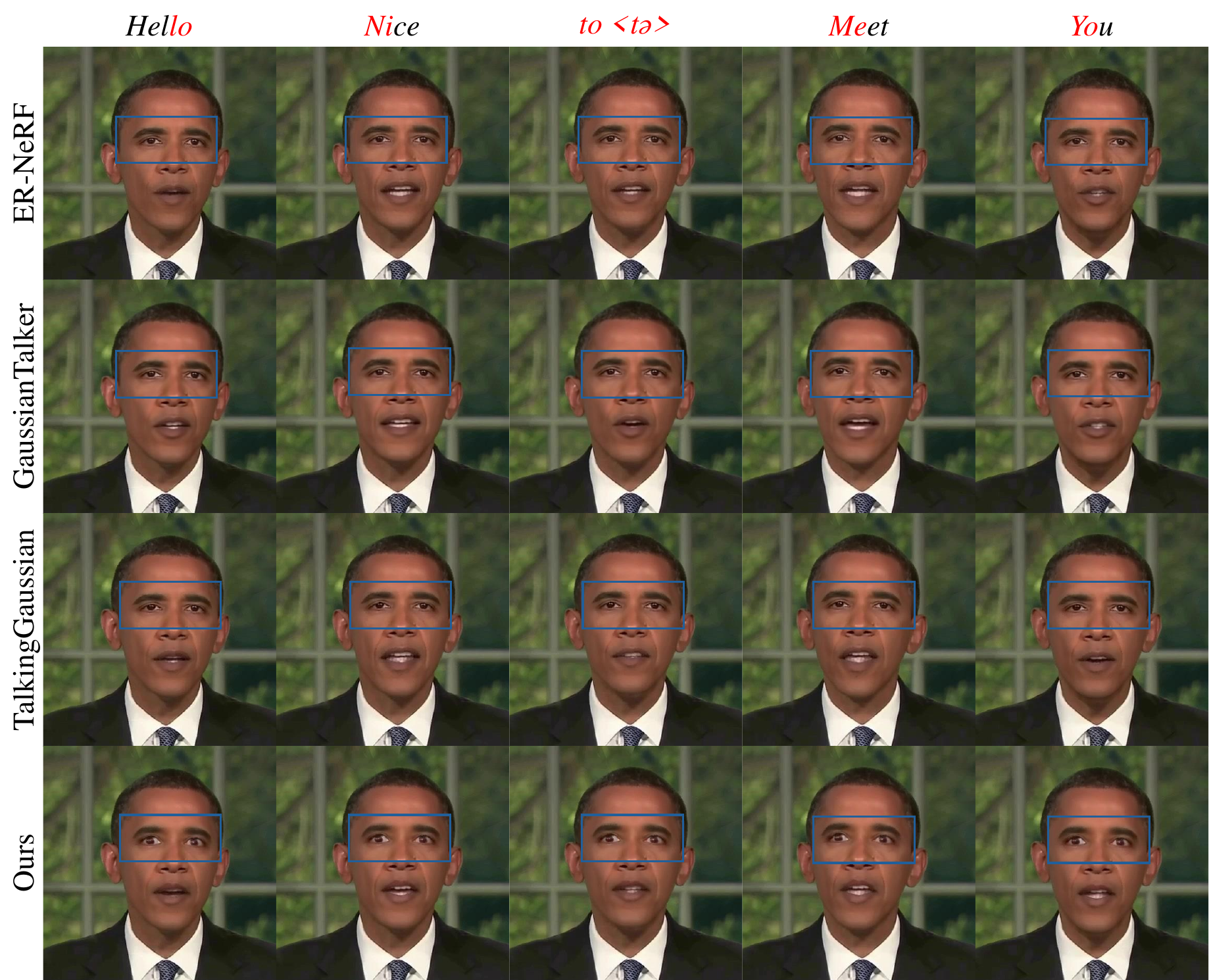}
    \caption{We demonstrate qualitative comparisons in the emotion-conditioned scenario against other methods, including ER-NeRF~\cite{li2023efficient}, GaussianTalker~\cite{cho2024gaussiantalker}, and TalkingGaussian~\cite{li2024talkinggaussian}. The valence and arousal values are set to 0.31 and 0.74, respectively, to represent the emotion of surprise. We use \textcolor{blue}{blue} boxes to highlight the area around the eyes, emphasizing the expression of emotion. The words being pronounced by the speaker in each column are highlighted in \textcolor{red}{red}.}
    \label{fig:scenario3}
\end{figure*}

\begin{figure*}[t]
    \centering
    \includegraphics[width=0.99\linewidth]{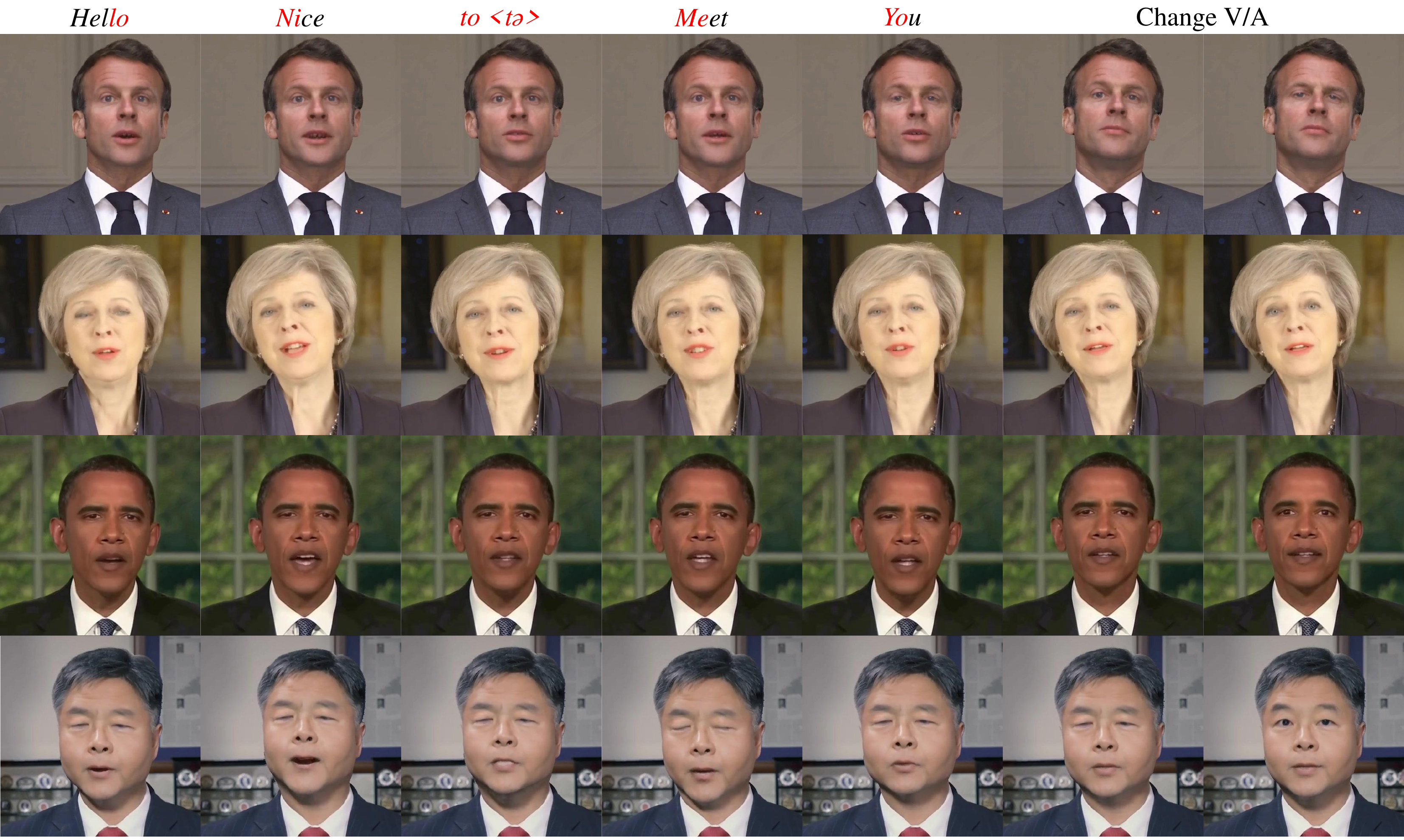}
    \caption{We demonstrate our proposed method's ability to reflect the desire emotion in the rendered face. Each row utilizes specific valence-arousal values: (0.3, 0.7), (0.8, -0.5), (-0.6, 0.5), and (-0.8, -0.2). The first row changes the valence/arousal from (0.3, 0.7) to (-0.8, 0.5). The second row changes the valence/arousal from (0.8, -0.5) to (0.6, 0.5). The third row changes the valence/arousal from (-0.6, 0.4) to (0.8, 0.2). The fourth row changes the valence/arousal from (-0.8, -0.2) to (0.3, 0.6). Additionally, the words being pronounced by the speaker in each column are highlighted in \textcolor{red}{red}.}
    \label{fig:change emotion}
\end{figure*}

We present the qualitative results across three scenarios: self-reconstruction, cross-domain audio, and emotion-conditioned generation, as shown in Figs.~\ref{fig:scenario1}, \ref{fig:scenario2}, and \ref{fig:scenario3}. For the each scenario, we compare our results with those of other models, including ER-NeRF~\cite{li2023efficient}, GaussianTalker~\cite{cho2024gaussiantalker}, and TalkingGaussian~\cite{li2024talkinggaussian}. In the emotion-conditioned scenario, we demonstrate the results using valence and arousal values of 0.31 and 0.74, respectively. Additionally, we illustrate our method's ability to reflect the desired emotion in the rendered face, as shown in Fig~\ref{fig:change emotion}. Furthermore, we showcase the transition of emotions by changing the valence and arousal values. For a detailed view of the continuous transitions in facial emotions and additional results, please refer to the supplementary video, including emotion-conditioned scenario comparison, valence-arousal interpolation, 360° valence-arousal interpolation (radius: 0.8), and dynamic emotion transitions during speech.

{
    \small
    \bibliographystyle{ieeenat_fullname}
    \bibliography{main}
}
\end{document}